\documentclass[journal]{IEEEtran}
\AtBeginDocument{%
  }

\usepackage{url}
\usepackage{flushend}
\usepackage{hyperref}
\hypersetup{
    colorlinks=true,
    linkcolor=blue,
    filecolor=blue,      
    urlcolor=blue,
    citecolor=cyan,
}
\usepackage{cite}
\usepackage{pifont}
\usepackage{booktabs}
\usepackage{xcolor}
\usepackage{colortbl}
\usepackage{color}
\usepackage{comment}
\usepackage{multirow}
\usepackage{graphicx}
\usepackage{subfigure}
\usepackage{textcomp}
\definecolor{mygreen}{RGB}{50,120,138}
\definecolor{myyellow}{RGB}{255,205,49}

\begin{document}
\title{FakeBench: Probing Explainable Fake Image Detection via Large Multimodal Models}
\author{Yixuan~Li, Xuelin Liu, Xiaoyang Wang, Bu Sung Lee,~\IEEEmembership{Senior Member,~IEEE}, Shiqi~Wang\IEEEauthorrefmark{2},~\IEEEmembership{Senior Member,~IEEE}, Anderson Rocha,~\IEEEmembership{Fellow,~IEEE}, and Weisi Lin,~\IEEEmembership{Fellow,~IEEE} 
\thanks{Y. Li, X. Liu, and S. Wang are with the Department of Computer and Science, City University of Hong Kong, Hong Kong (e-mail: yixuanli423@gmail.com; xuelinliu-bill@foxmail.com; shiqwang@cityu.edu.hk). 

X. Wang is with the College of Computing and Informatics, Drexel University, USA (e-mail: xw388@drexel.edu).

A. Rocha is with the Artiﬁcial Intelligence Lab. (Recod.ai) at the University of Campinas, Campinas 13084-851, Brazil (e-mail: arrocha@unicamp.br)

B.S. Lee and W. Lin are with the College of Computing and Data Science, Nanyang Technological University, Singapore (e-mail: ebslee@ntu.edu.sg; wslin@ntu.edu.sg).

\IEEEauthorrefmark{2}Corresponding author: Shiqi Wang.

This paper has supplementary materials provided by the authors, including additional experimental details and analyses.}
}
\markboth{}%
{Shell \MakeLowercase{\textit{et al.}}: Bare Demo of IEEEtran.cls for IEEE Communications Society Journals}
\maketitle
\begin{abstract}
The ability to distinguish whether an image is generated by artificial intelligence (AI) is a crucial ingredient in human intelligence, usually accompanied by a complex and dialectical forensic and reasoning process. However, current fake image detection models and databases focus on binary classification without understandable explanations for the general populace. This weakens the credibility of authenticity judgment and may conceal potential model biases. Meanwhile, large multimodal models (LMMs) have exhibited immense visual-text capabilities on various tasks, bringing the potential for explainable fake image detection. Therefore, we pioneer the probe of LMMs for explainable fake image detection by presenting a multimodal database encompassing textual authenticity descriptions, the FakeBench. For construction, we first introduce a fine-grained taxonomy of generative visual forgery concerning human perception, based on which we collect forgery descriptions in human natural language with a human-in-the-loop strategy. FakeBench examines LMMs with four evaluation criteria: \textit{detection}, \textit{reasoning}, \textit{interpreting} and \textit{fine-grained forgery analysis}, to obtain deeper insights on image authenticity-relevant capabilities. Experiments on various LMMs confirm their merits and demerits in different aspects of fake image detection tasks. This research presents a paradigm shift towards \textit{transparency} for the fake image detection area and reveals the need for greater emphasis on forensic elements in visual-language research and AI risk control. FakeBench will be available at \url{https://github.com/Yixuan423/FakeBench}.
\end{abstract}

\begin{IEEEkeywords}
Large multimodal models, fake image detection, explainability, benchmark, image forensics
\end{IEEEkeywords}
\maketitle

\section{Introduction}
\begin{figure*}[!t]
    \includegraphics[scale=0.87]{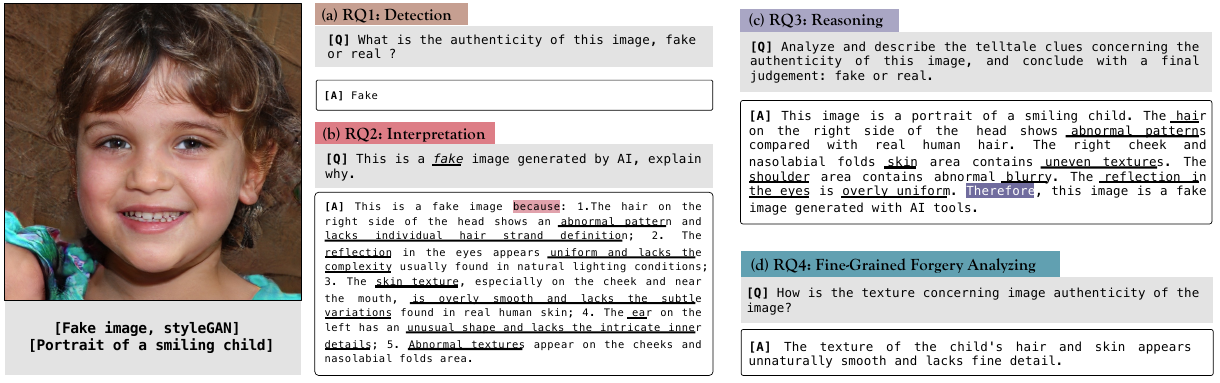}
    \centering
    \caption{Scope of research questions with exemplar question-answer pairs from the proposed FakeBench dataset. (a) Fake image detection; (b) fake image reasoning; (c) fake image interpretation; (d) fine-grained forgery analyses. The responses are exemplars provided by humans.}
    \label{fig:rqs}
\end{figure*}
\IEEEPARstart{T}he rapid advancement of generative artificial intelligence (AI)~\cite{midjourney,betker2023improving} escalates the threats to cybersecurity, such as disinformation campaigns and swindling, which are particularly pronounced in the realm of AI-synthetic fake images. Recently, substantial efforts have been devoted to counter the abuses by advancing fake image detection\footnote{In the context of this paper, we refer ``fake images'' in particular to the image synthesized by generative AI models with generic contents. We use ``image defaking'' and ``fake image detection'' alternately.} algorithms~\cite{barni2020cnn,frank2020leveraging,gragnaniello2021gan,ju2022fusing,liu2020global,liu2022detecting,tan2023learning}. Normally, image defaking techniques aim to identify fakes and manipulations by analyzing image patterns that may indicate forgery. However, restricting image defaking to the binary classification scenario has increasingly failed to earn people's trust, especially as the sophistication of fake images continues to improve~\cite{tariang2024synthetic}. This contradiction underscores the need for a more convincing approach beyond classification: \textit{explanations} with detailed evidence, thereby validating outputs and revealing potential model biases. 

Existing explainable image defaking methods typically provide latent representations\cite{dong2022explaining,AghasanliICCVW2023}, feature illustrations~\cite{chai2020makes,bird2024cifake}, and artifact localization~\cite{zhang2023perceptual} in addition to binary authenticity judgments to enhance interpretability. Nonetheless, such abstract signs face challenges in explaining the underlying decision-making process, particularly regarding very nuanced forgery cues and fine-grained flaw analyses. Recently, Zhang et al.~\cite{zhang2024common} and Sun et al.~\cite{sun2023towards} have taken the lead in investigating the explainability of deepfake face detection. They utilized successively connected unimodal encoders to unify modals and achieved promising results, advancing interpretable defaking to a new stage. Therefore, it is crucial to further advance fake image detection from a black-box approach to a more \textbf{transparent} one. In this paper, we rely on one of the most intuitive and flexible mediums, human natural language, towards explainability for image defaking. While it is of great challenge for conventional unimodal defaking models~\cite{wang2020cnn,WangICCV2023,ZhongArxiv2023,hou2023evading,ojha2023towards,xu2023exposing} to do so, the recent development of large multimodal models (LMMs) shed light on \textit{flexible}, \textit{interpretable} and \textit{text-driven} fake image detection in a novel perspective.

As the collaborative evolution of Natural Language Processing (NLP) and Computer Vision (CV), LMMs have marked a significant milestone in AI~\cite{liu2024visual,bai2023QwenVL,dai2023instructblip}. LMMs integrate visual and textual data via the connection and alignment of Large Language Models (LLMs) and Vision Transformers (ViTs)~\cite{dosovitskiy2020image}, exhibiting impressive high-level visual reasoning and low-level vision~\cite{liu2023mmbench,ge2024openagi,zhu20242afc}. LMMs are unique in humanoid comprehension, extensive exposure to knowledge, text-driven evaluations, flexible query formats, and human-accessible contextual explanations~\cite{zhao2024explainability} compared with unimodal frameworks. Ideally, LMMs are expected to benefit image defaking in the following aspects: recognizing and describing visual forgery cues, flexible text-conditioned analyses, and causal investigation for authenticity. As such, we present a paradigm shift for fake image detection with LMMs through what we call \textit{transparent defaking}. This LMM-based approach emphasizes both image authenticity classification and the process behind the judgment. However, the authenticity-relevant aspects of LMMs still remain unexplored, and the primary obstacle is the lack of natural language annotations concerning visual forgery cues for fake images. 

To fill this void, we introduce \textbf{FakeBench}, a pioneering multimodal dataset probing LMMs on transparent image defaking. FakeBench encompasses detailed long textual authenticity descriptions, emphasizing \textit{visual characteristics}, \textit{semantic inconsistencies}, and \textit{commonsense violations} to characterize fakery. Furthermore, it features multi-dimensional evaluations, including simple detection, causal investigations, and fine-grained forgery analysis, utilizing a variety of NLP prompting strategies (\textit{i.e.}, basic, in-context, chain-of-thought, and free prompting) to probe LMMs. Specifically, FakeBench focuses on the following research questions (RQs).
\begin{itemize}
    \item \textbf{RQ1.} Can LMMs {\color{mygreen}\textbf{\textit{detect}}} fake images generated by various generative models? As shown in Fig.~\ref{fig:rqs}(a), LMMs are expected to distinguish fake and real images apart, \emph{e.g.}, answering ``\textit{Fake}'' for AI-generated images when quizzed with `\textit{What is the authenticity of this image?}'
    \item \textbf{RQ2.} Can LMMs {\color{mygreen}\textbf{\textit{interpret}}} the detection result adequately? As depicted in Fig.~\ref{fig:rqs}(b), LMMs are expected to give the forgery evidence supporting the given authenticity, emphasizing effect-to-cause projection from image authenticity to visual forgery clues.
    \item \textbf{RQ3.} Can LMMs perform complex {\color{mygreen}\textbf{\textit{reasoning}}} on fake images? As illustrated in Fig.~\ref{fig:rqs}(c), LMMs are expected to provide logical chain-of-thought (CoT) reasoning via natural language before reaching the final judgment, emphasizing the cause-to-effect projection from visual forgery to image authenticity.
    \item \textbf{RQ4.} Can LMMs {\color{mygreen}\textbf{\textit{analyze}}} fine-grained forgery aspects? As shown in Fig.~\ref{fig:rqs}(d), models shall discuss certain aspects regarding image authenticity.
\end{itemize}
Guided by the RQs, FakeBench is designed to encompass three ingredients. First, we establish the \textbf{FakeClass} dataset with 6,000 diverse-sourced fake and genuine images accompanied by Question\&Answers ($\mathtt{Q}$\&$\mathtt{A}$) on image authenticity judgment (RQ1). Second, we construct the \textbf{FakeClue} dataset containing meticulous descriptions of forgery cues for 3,000 fake images, which are reliably and efficiently collected with a human-in-the-loop strategy. In FakeClue, we propose two mutual-reverse text-driven strategies in accord with human causal thoughts, including fault-finding mode (RQ2) and inference mode (RQ3). Third, we propose \textbf{FakeQA} containing 42,000 open-ended $\mathtt{Q\&A}$ pairs to investigate whether LMMs can analyze fine-grained forgery aspects (RQ4). Based on FakeBench, we provide extensive evaluations with 14 foundational LMMs and offer some intriguing insights. We discovered that some LMMs have demonstrated zero-shot detection capabilities close to human intelligence and even surpass specialized models. However, their reasoning and interpretation abilities still have room for improvement, even for advanced proprietary ones. Besides, CoT prompting demonstrates unusual inefficacy in detection due to the lack of authenticity-relevant concepts and the deficient projection from forgery descriptions to judgments in general-purpose LMMs.

In summary, this paper presents a heuristic and systematic exploration of explainable fake image detection with LMMs. We hope \textbf{FakeBench} to inspire the community's ongoing efforts to achieve transparent image defaking. Our primary contributions are threefold.
\begin{itemize}
    \item We establish the first multimodal dataset on explainable image authenticity assessment, probing LMM's capabilities of \textit{simple detection, interpretation, reasoning}, and \textit{fine-grained forgery analysis}.
    \item We present a fine-grained taxonomy for generative forgery cues to guide \textit{what to explain} in transparent defaking and calibrate \textit{how far} LMMs are from humans in the tug-of-war between image generation and detection.
    \item We conduct in-depth experimental analyses on 14 LMMs and draw some findings on performance, inter-task relevance, chain-of-thought prompting influence, and the effect of fine-tuning data. These offer deeper insights into the merits and demerits of foundational LMMs on explainable fake image detection for future explorations.
\end{itemize}

\section{Related Work}
\subsection{AI-Synthetic Image Detection}
To mitigate the impacts of the abuse of AI-synthetic images, numerous attempts at fake image detection have been proposed during recent years~\cite{lin2024detecting,sha2023fake,wu2023cheap,chang2023antifakeprompt}. The attempts have been focusing on the generalization ability~\cite{ju2022fusing,tan2023learning,ojha2023towards,zhu2023gendet}, robustness~\cite{BorjiIVC2023,WangICCV2023,ZhongArxiv2023,WolterML2023}, and generator attribution~\cite{chai2020makes,sha2023fake, EpsteinICCVW2023,AghasanliICCVW2023,JUTMM2024}. However, current models generally classify genuine or fake images for a binary output, which is opaque for human comprehension. This characteristic risks diminishing the reliability and verifiability of the model's judgments on image authenticity. On the other hand, AI-synthetic images often feature generative artifacts such as unnatural hues and fault perspectives due to the unique architecture of generation models~\cite{tariang2024synthetic}. Such semantic inconsistencies are usually salient and sometimes subtle, while current generative models cannot completely avoid them. Even powerful diffusion models possess systematic shadow and perspective errors~\cite{sarkar2024shadows}. When properly captured and conveyed, humans can comprehend these perceptible traces well, making them an appropriate demonstration for the model's judgment.

Many well-established datasets for image defaking have been proposed to handle the ever-increasing difficulty~\cite{wang2019fakespotter,wang2020cnn,verdoliva2022}. The emphases of recent works are on improving data scale~\cite{he2021forgerynet,lu2024seeing,zhu2024genimage}, content diversity~\cite{zhu2024genimage,wang2023benchmarking,chang2023antifakeprompt}, and generator diversity~\cite{lu2024seeing,chang2023antifakeprompt}. The comparisons with previous datasets for AI-generated image detection are listed in Table~\ref{dbcompare}. It is shown that the proposed FakeBench is the first fine-grained multimodal dataset on explainable fake image detection dedicated to multiple subtasks simultaneously. 


\begin{table}[tbp]
\caption{Comparisons with existing fake image detection datasets. The Diff. and AutoReg. are short for Diffusion and Autoregressive Models, and $\ast$ indicates non-public datasets.}
\centering
\fontsize{6}{7}\selectfont
\renewcommand\arraystretch{1}
\setlength{\tabcolsep}{0.5mm}{
\begin{tabular}{lcccccccc}
\toprule[0.7pt]
\multirow{2}{*}{Dataset} & \multirow{2}{*}{Content} &\multirow{2}{*}{\#Generator} & \multicolumn{3}{c}{Generator Type}       & \multirow{2}{*}{Fine-Grained}  &\multirow{2}{*}{Multimodal}&\multirow{2}{*}{Multitask} \\ \cline{4-6}
&               & & \textit{GAN}        & \textit{Diff.} & \textit{AutoReg.} &                         &                                     \\ \hline

FakeSpotter~\cite{wang2019fakespotter}  & Face & 7 & \color[HTML]{036400}{\ding{52}} & \color[HTML]{CB0000}{\ding{56}}          & \color[HTML]{CB0000}{\ding{56}}  & \color[HTML]{CB0000}{\ding{56}}  &\color[HTML]{CB0000}{\ding{56}} &\color[HTML]{CB0000}{\ding{56}}  \\
ForgeryNet~\cite{he2021forgerynet}   & Face   & 15 & \color[HTML]{036400}{\ding{52}}          & \color[HTML]{CB0000}{\ding{56}}          & \color[HTML]{CB0000}{\ding{56}}              & \color[HTML]{036400}{\ding{52}}            &\color[HTML]{CB0000}{\ding{56}} &\color[HTML]{CB0000}{\ding{56}}  \\
DeepArt~\cite{wang2023benchmarking}                                & Art                       & 5      & \color[HTML]{CB0000}{\ding{56}}          & \color[HTML]{036400}{\ding{52}}          & \color[HTML]{CB0000}{\ding{56}}              & \color[HTML]{CB0000}{\ding{56}}               &\color[HTML]{CB0000}{\ding{56}} &\color[HTML]{CB0000}{\ding{56}}    \\
CNNSpot~\cite{wang2020cnn} & Object   & 11   & \color[HTML]{036400}{\ding{52}}   & \color[HTML]{CB0000}{\ding{56}}          & \color[HTML]{CB0000}{\ding{56}}              & \color[HTML]{CB0000}{\ding{56}}             &\color[HTML]{CB0000}{\ding{56}} &\color[HTML]{CB0000}{\ding{56}}     \\
IEEE VIP Cup~\cite{verdoliva2022}  & Object  & 5  & \color[HTML]{036400}{\ding{52}}          & \color[HTML]{036400}{\ding{52}}          & \color[HTML]{CB0000}{\ding{56}}                & \color[HTML]{CB0000}{\ding{56}}   &\color[HTML]{CB0000}{\ding{56}} &\color[HTML]{CB0000}{\ding{56}}  \\
CiFAKE~\cite{bird2024cifake}   & Object & 1  & \color[HTML]{CB0000}{\ding{56}}          & \color[HTML]{036400}{\ding{52}}          & \color[HTML]{CB0000}{\ding{56}}    & \color[HTML]{CB0000}{\ding{56}}                 &\color[HTML]{CB0000}{\ding{56}} &\color[HTML]{CB0000}{\ding{56}}  \\
UniversalFake~\cite{ojha2023towards} &General &4 &\color[HTML]{CB0000}{\ding{56}}& \color[HTML]{036400}{\ding{52}} & \color[HTML]{036400}{\ding{52}} &\color[HTML]{CB0000}{\ding{56}}&\color[HTML]{CB0000}{\ding{56}}&\color[HTML]{CB0000}{\ding{56}}\\
DE-FAKE$\ast$~\cite{sha2023fake} &General &4&\color[HTML]{CB0000}{\ding{56}}&\color[HTML]{036400}{\ding{52}} &\color[HTML]{036400}{\ding{52}} &\color[HTML]{CB0000}{\ding{56}} &\color[HTML]{036400}{\ding{52}}  &\color[HTML]{CB0000}{\ding{56}} \\
GenImage~\cite{zhu2024genimage} &General &8  &\color[HTML]{036400}{\ding{52}}            &  \color[HTML]{036400}{\ding{52}}&\color[HTML]{CB0000}{\ding{56}}            &  \color[HTML]{CB0000}{\ding{56}}             &  \color[HTML]{CB0000}{\ding{56}}   &\color[HTML]{CB0000}{\ding{56}} \\
AntifakePrompt~\cite{chang2023antifakeprompt} &General &5 &\color[HTML]{CB0000}{\ding{56}}&\color[HTML]{036400}{\ding{52}}&\color[HTML]{CB0000}{\ding{56}} &\color[HTML]{CB0000}{\ding{56}}&\color[HTML]{036400}{\ding{52}}&\color[HTML]{CB0000}{\ding{56}}\\
Fake2M~\cite{lu2024seeing} &General &11 &\color[HTML]{036400}{\ding{52}}&\color[HTML]{036400}{\ding{52}}&\color[HTML]{036400}{\ding{52}}&\color[HTML]{CB0000}{\ding{56}}&\color[HTML]{CB0000}{\ding{56}}&\color[HTML]{CB0000}{\ding{56}}\\
WildFake~\cite{hong2024wildfake} &General &21 &\color[HTML]{036400}{\ding{52}} &\color[HTML]{036400}{\ding{52}} &\color[HTML]{036400}{\ding{52}} &\color[HTML]{CB0000}{\ding{56}}&\color[HTML]{CB0000}{\ding{56}}&\color[HTML]{CB0000}{\ding{56}}\\
\rowcolor{gray!15}
\textbf{FakeBench}      & General  &10           & \color[HTML]{036400}{\ding{52}} & \color[HTML]{036400}{\ding{52}} & \color[HTML]{036400}{\ding{52}}     & \color[HTML]{036400}{\ding{52}}             &  \color[HTML]{036400}{\ding{52}} &\color[HTML]{036400}{\ding{52}}                     \\ 
\bottomrule[0.7pt]
\end{tabular}
}
\label{dbcompare}
\end{table}
\begin{table}[!t]
    \centering
    \fontsize{7}{9}\selectfont
    \renewcommand\arraystretch{1.1}
    \caption{Overview of the source of AI-generated fake images in the FakeBench. The type size is reflected by the number of images.}
    \setlength{\tabcolsep}{0.4mm}{\begin{tabular}{lllc}
    \toprule
    Type &Base Model & Source Dataset & Type Size\\
    \midrule
        ProGAN~\cite{karras2017progressive} &GAN &CNNSpot~\cite{wang2020cnn} &300\\
    \arrayrulecolor{gray!25}
    \midrule
    \arrayrulecolor{black}
        StyleGANs~\cite{karras2019style} &GAN &CNNSpot~\cite{wang2020cnn} &300\\
    \midrule
        CogView2~\cite{ding2022cogview2} &AutoRegressive &HPS v2~\cite{wu2023human} &300\\
    \midrule
    \arrayrulecolor{gray!25}
        FuseDream~\cite{liu2021fusedream} &Diffusion Model &HPS v2~\cite{wu2023human} &300\\
    \midrule
        VQDM~\cite{gu2022vector} &  Diffusion Model&GenImage~\cite{zhu2024genimage} &300\\
    \midrule
        \multirow{2}[1]{*}{Glide~\cite{nichol2021glide}} & \multirow{2}[1]{*}{Diffusion Model} &HPS v2~\cite{wu2023human} &\multirow{2}[1]{*}{300}\\
        & &GenImage~\cite{zhu2024genimage} \\
    \midrule
    \arrayrulecolor{black}
        \multirow{4}[1]{*}{Stable Diffusion (SD)~\cite{rombach2022high}} & \multirow{4}[1]{*}{Diffusion Models}&AGIQA-3K~\cite{li2023agiqa} &\multirow{4}[2]{*}{300}\\
        & &I2IQA~\cite{yuan2023pku} \\
        & &NIGHTS~\cite{fu2023dreamsim} \\
        &&DiffusionDB~\cite{wang2022diffusiondb}\\ 
    \midrule
    \multirow{2}[1]{*}{DALL$\cdot$E2~\cite{ramesh2022hierarchical}} &\multirow{2}[1]{*}{Proprietary}  &AGIQA-3K~\cite{li2023agiqa} &\multirow{2}[2]{*}{300}\\
    & &HPS v2~\cite{wu2023human}\\
    \arrayrulecolor{gray!25}
    \midrule
        \multirow{2}[2]{*}{DALL$\cdot$E3~\cite{betker2023improving}} &\multirow{2}[1]{*}{Proprietary} &Dalle3-reddit-dataset~\cite{dalle3reddit} &\multirow{2}[2]{*}{300}\\
        & &DALL$\cdot$E3 dataset~\cite{betker2023improving}\\
    \arrayrulecolor{black}
    \midrule
        \multirow{3}[2]{*}{Midjourney (MJ)~\cite{midjourney}} &\multirow{3}[2]{*}{Proprietary} &AGIQA-3K~\cite{li2023agiqa} &\multirow{3}[2]{*}{300}\\
        & &I2IQA~\cite{yuan2023pku}\\
        & &Midjourney-v5~\cite{midjourney-v5-dataset} \\
    \bottomrule
    \end{tabular}%
    }
\label{tab:imagesource}
\end{table}
\begin{figure}[!tbp]
    \centering  
    \includegraphics[scale=0.5]{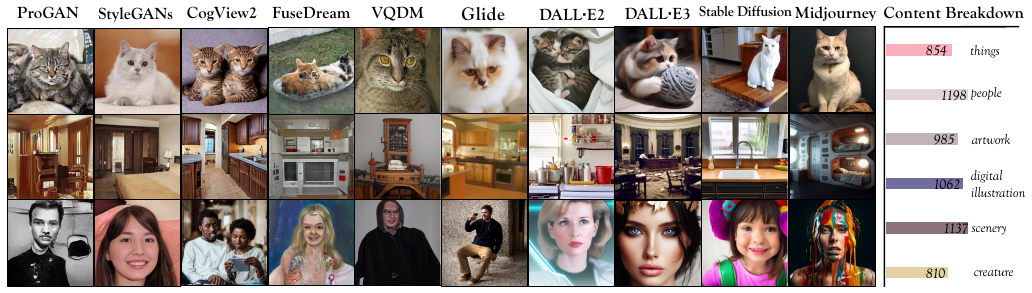}
    \caption{Examples and content breakdown for the fake images in FakeBench. In total, ten different deep generative models are incorporated with balanced content.}
    \label{fig:imgcontent}
\end{figure}
\subsection{Large Multimodal Models}
The evolution of LMMs has marked a significant milestone in AI, which provides impressive visual capabilities and seamless interaction ability with humans through natural languages. Open-source LMMs~\cite{liu2024visual,zhu2023minigpt,dai2023instructblip,li2023otter} and commercial ones such as GPT-4V~\cite{achiam2023gpt}, GeminiPro~\cite{geminiteam2024gemini}, and Claude3~\cite{claude3} have exhibited remarkable abilities in tackling various multi-modality tasks. LMMs have been validated by diversified benchmarks to see if they are all-around players~\cite{li2024seed,liu2023mmbench,cai2023benchlmm,shi2023chef} or on individual vision-language tasks such as image quality assessment~\cite{wu2024comprehensive}, image aesthetics perception~\cite{huang2024aesbench}, science question answering~\cite{lu2022learn}, and commonsense reasoning~\cite{talmor2018commonsenseqa}. Nevertheless, the ability of LMMs to detect image defaking remains undetermined. Therefore, we construct the first multimodal fake image detection dataset with explanations in natural languages, the \textbf{FakeBench}. We are concerned about whether LMMs can distinguish between real and fake and the underlying process beneath authenticity judgments to comprehensively evaluate their multi-dimensional ability on transparent image defaking.

\section{FakeBench Construction}
In this section, we present the construction of FakeBench, including guiding rationale, image collection, generative forgery taxonomy, and three FakeBench ingredients for different aspects of defaking capabilities.
\subsection{Guiding Rationale}
\label{sec:guidingthoughts}
The proposed FakeBench is constructed upon three rationales: \textbf{1)} \textbf{Focusing on multi-dimensional abilities related to transparent image defaking.} Unlike conventional image defaking, which typically involves simple binary classification, FakeBench employs a holistic approach. It integrates LMM-based \textit{detection}, effect-to-cause \textit{interpretation}, cause-to-effect \textit{reasoning}, and \textit{fine-grained forgery analysis} simultaneously to thoroughly explore LMM capabilities in image defaking. \textbf{2)} \textbf{Covering diverse generative visual forgery orchestrating with various image content.} FakeBench is designed to encompass a broad spectrum of forgery appearances generated by various image generation models. This ensures a rich diversity in the image content and generative patterns included in the benchmark. \textbf{3)} \textbf{Ensuring that descriptive annotations on generative forgery resonate with human understanding.} To effectively evince the fakes through forgery descriptions, figuring out what influences human perceptions of authenticity is essential. Therefore, a human-participated subjective study is crucial to determine \textit{what to explain}, guiding the creation of informative and relevant descriptions.
\subsection{Image Preparation}
\label{sec:imageprepare}
To cater to increasingly diversified image generators, the generation models included in the FakeBench span ten different types, including two generative adversarial networks (GAN) based models~\cite{karras2017progressive,karras2019style}, one auto-regressive model~\cite{ding2022cogview2}, four stable diffusion (SD) based models~\cite{liu2021fusedream, gu2022vector, nichol2021glide,rombach2022high}, and three proprietary models~\cite{ramesh2022hierarchical,betker2023improving,midjourney}. The details are listed in Table~\ref{tab:imagesource}. To enhance the diversity of fake pattern and image content, the fake images are sampled from multiple datasets (see Table~\ref{tab:imagesource}), including CNNSpot~\cite{wang2020cnn}, HPS v2~\cite{wu2023human}, AGIQA-3K~\cite{li2023agiqa}, GenImage~\cite{zhu2024genimage}, NIGHTS~\cite{fu2023dreamsim}, I2IQA~\cite{yuan2023pku}, DiffusionDB~\cite{wang2022diffusiondb}, DALL$\cdot$E3 dataset~\cite{betker2023improving}, Dalle3-reddit-dataset~\cite{dalle3reddit}, and Midjourney-v5 dataset~\cite{midjourney-v5-dataset}. Regarding the genuine images, we sample 3,000 images from ImageNet~\cite{deng2009imagenet} and DIV2K dataset~\cite{agustsson2017ntire}. The AI-generated images are carefully chosen to ensure content diversification, which includes \textit{things}, \textit{people}, \textit{artwork}, \textit{digital illustration}, \textit{scenery}, and \textit{creature}. Each image is assigned \textit{at least one} content label. The examples for the image content of FakeBench are illustrated in Fig.~\ref{fig:imgcontent}.
\subsection{Fine-Grained Generative Visual Forgery Taxonomy}
\label{sec:forgerytaxonomy}
The factors influencing human judgments of image authenticity are still undetermined and complex questions~\cite{park2024can}. How to characterize generative forgery cues from the visual perspective needs to be clarified to align with human comprehension. Relevant works either draw a rough equivalence between authenticity and low-level image characteristics~\cite{chen2023exploring} or focus on certain aspects such as content smudging and false perspective~\cite{lu2024seeing,he2021forgerynet,farid2022lighting,farid2022perspective}. To remedy this issue, we construct a comprehensive taxonomy of visual forgery cues for AI-generated images and validate it with human-participated subjective studies. Our objective is twofold: 1) benchmarking human distinguishing ability towards real and fake images, and 2) focalizing human visual criteria in image authenticity judgment. The results can function as the baseline to benchmark LMM's humanoid intelligence towards image defaking and guide the description of generative forgery.

\begin{figure}[!tbp]
    \centering  
    \includegraphics[scale=0.43]{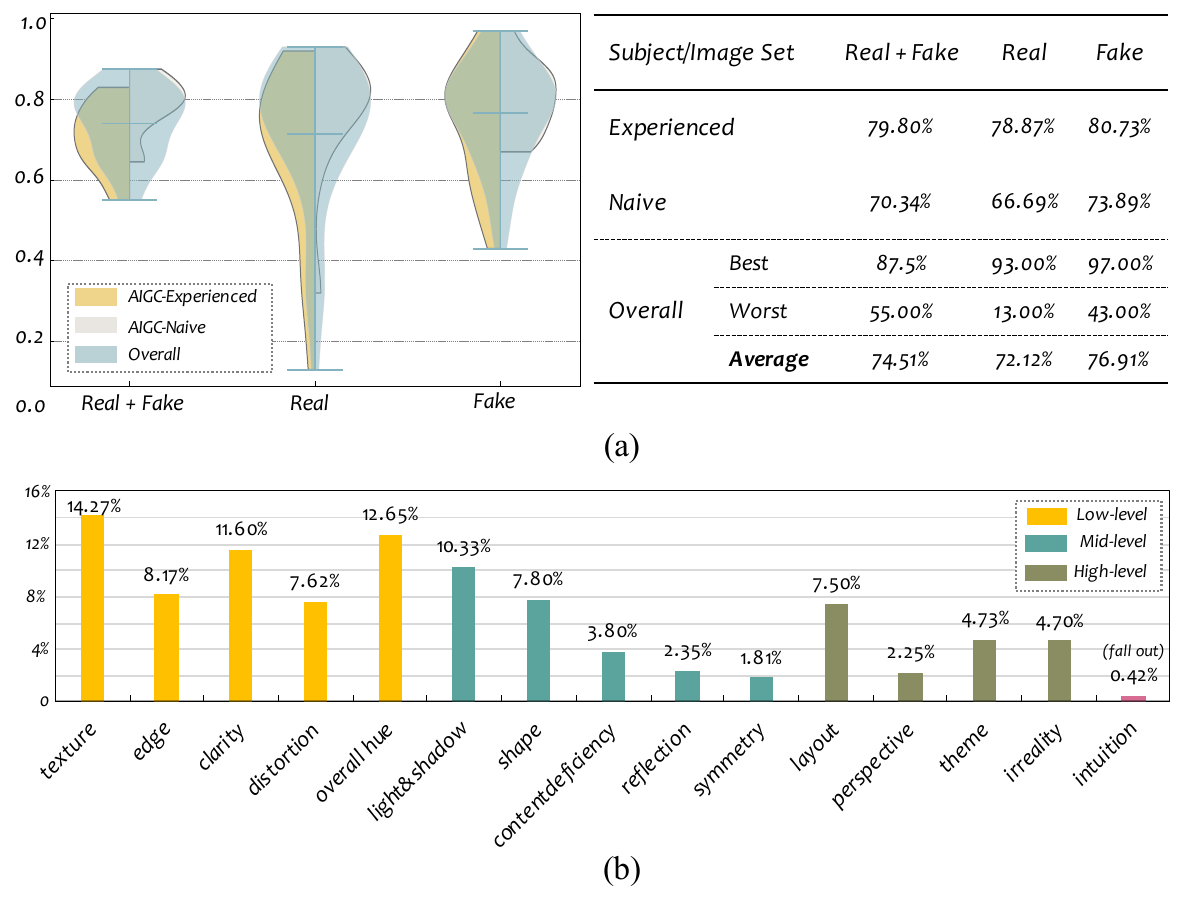}
    \caption{Subjective analyses on image authenticity judgment. (a) Human distinguishing analyses regarding accuracy on genuine and fake images. The left part illustrates the estimated distributions for the accuracy of each authenticity category, and the detailed statistics are on the right side. (b) The proposed taxonomy of authenticity-relevant judging criteria and human-rating statistics. Different human subjects contribute to each subset's best and worst performances.}
    \label{fig:subj}
\end{figure}

The subjective study is conducted on genuine images and AI-generated images with diversified contents, generators, and fine-grained visual forgery aspects. We evenly sampled 200 fake and 200 genuine images from Sec.~\ref{sec:imageprepare} and recruited \textbf{34} participants of balanced backgrounds (17 naive and 17 experienced) in image generation to ensure statistical significance, fairness, and reliability. Each participant is asked to distinguish whether each image is generated by AI with no time limit and then select no less than one forgery aspect evincing their judgments. The candidate forgery types are the preset 14 criteria influencing the authenticity-relevant judgment of humans, making up the taxonomy of generative visual forgery cues. According to the level of abstraction, we consider five low-level criteria including \textit{texture, edge, clarity, distortion, and overall hue}, five mid-level criteria including \textit{light\&shadow, shape, content deficiency, symmetry, and reflection}, and four high-level vision criteria including \textit{layout, perspective, theme, and irreality}. In practice, we further provided one fallback option ``\textit{intuition}'' for the occasion where none of the preset forgery satisfy perception. Classification accuracy (\textit{ACC}) functions as the measure. The averaged performance and the number of each criterion for correctly distinguished images are shown in Fig.~\ref{fig:subj}(a) and Fig.~\ref{fig:subj}(b), respectively. Detailed textual definitions and illustrations of forgery cues are provided in the supplementary material.

We discover some patterns concerning human distinguishing abilities from Fig.~\ref{fig:subj}(a). Humans obtain an \textbf{average accuracy of 74.51\%} on assessing image authenticity, and the \textbf{peak performance reaches 87.50\%}. However, human performance diversifies in terms of knowledge background and image authenticity, where experience in AI-generated content (AIGC) lowers the minimum accuracy but generally increases the mean accuracy. Additionally, all participants demonstrate lower average accuracy on real images than fake ones, indicating AIGC significantly erodes human trust, even when exposed to genuine information.
Moreover, Fig.~\ref{fig:subj}(b) indicates that the fallback option ``\textit{intuition}'' merely occupies 0.42\% of the total number of selections, meaning the other \textbf{99.58\%} selections of successfully-identified images correspond to at least one of the predefined forgery criteria. This validates the proposed taxonomy shows good coverage of visual forgery cues comprehended by humans, allowing to guide the subsequent description formulation on telltale clues in Sec.~\ref{sec:fakebench} reliably. 
\subsection{FakeBench Ingredients}
\label{sec:fakebench}
Considering the text-driven nature of LMMs, FakeBench formulates the explainable image defaking tasks in the form of visual question answering (VQA). We present four different prompting modes in correspondence to LMM's abilities in detection, interpretation and reasoning, and forgery analysis, which are encompassed by the three ingredients of FakeBench, including FakeClass, FakeClue, and FakeQA. 
\subsubsection{FakeClass}
\label{sec:fakeclass}
\begin{figure}[!t]
    \centering  
    \includegraphics[scale=0.95]{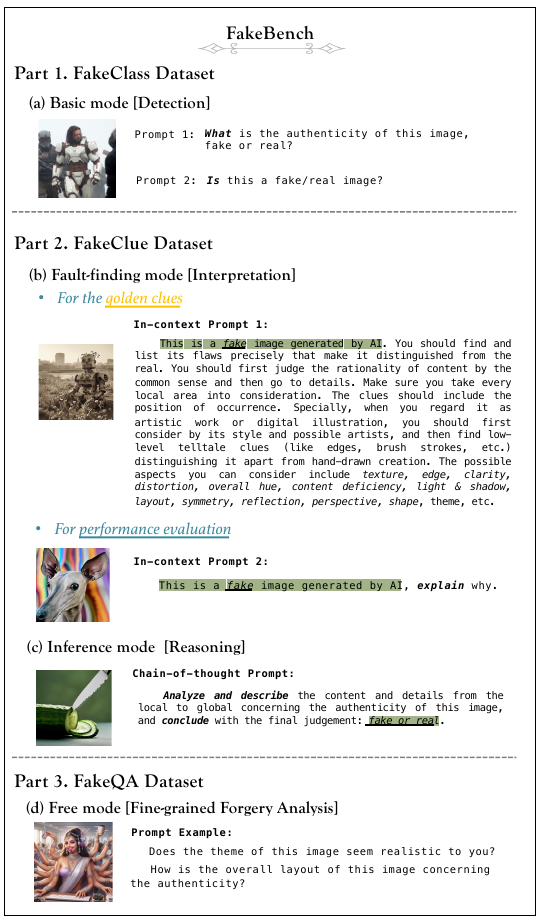}
    \caption{Prompt sheet of the proposed \textbf{FakeBench}, including: (a) basic mode for \textbf{FakeClass}, (b) fault-finding mode based on in-context prompting for \textbf{FakeClue}, (c) inference mode based on CoT prompting for \textbf{FakeClue}, and (d) free mode for \textbf{FakeQA}.}
    \label{fig:overview}
\end{figure}

In the first part of FakeBench, we evaluate the fake image {\color{mygreen}\textbf{\textit{detection}}} ability of LMMs, \emph{i.e.}, whether they can answer simple queries concerning objective image authenticity in natural languages. Therefore, for each fake image $\mathtt{I_0}$ and real image $\mathtt{I_1}$, we collect one authenticity-related question ($\mathtt{Q}$) in the closed-ended form, which is called \textit{basic mode} prompting. In particular, $\mathtt{Q}$s are in two different forms: `\textit{Yes-or-No}' and `\textit{What}' questions (see Fig.~\ref{fig:overview}(a)), \emph{i.e.}, `\textit{Is this a fake/real image?}' and `\textit{What is the authenticity of this image?}', which simulate common ways of human inquiry endow flexible evaluation approaches. The number of questions in different formats is balanced. Accordingly, we provide one correct answer ($\mathtt{A}$) to each question, which is in one of the forms of ``\textit{yes}'', ``\textit{no}'', ``\textit{real}'', and ``\textit{fake}'' to reflect the real ability of fake image detection. The 6,000 pieces of $\mathtt{(I_0,{Q}_{i},A)}$ tuples constitute \textbf{FakeClass}, where $i=0$ denotes the \textit{basic mode}.

\noindent{\textit{\textbf{Evaluation}}}\quad
The fake image detection ability of models is evaluated with the \textit{basic mode} prompting, and the answers for the closed-ended questions are compared by exact matching, \textit{i.e.}, the specific keyword concerning authenticity in responses must be identical to answers. The responses like ``\textit{fake}'' or ``\textit{I cannot answer yes or no.}'' are wrong when queried with ``\textit{Is this a fake image?}'', while the conditional existence of ``\textit{yes}'' shall be regarded as correct answers. 
\subsubsection{FakeClue}
\label{sec:fakeclue}
\begin{figure}[!tbp]
    \centering  
    \includegraphics[scale=1.1]{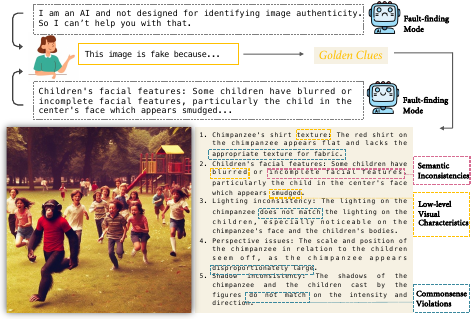}
    \caption{Human-in-the-loop strategy for producing the golden clues. Human and GPT-4V work in collaboration to ensure the quality of descriptions.}
    \label{fig:goldenclues}
\end{figure}
For the second part of FakeBench, we evaluate the {\color{mygreen}\textbf{\textit{reasoning}}} and {\color{mygreen}\textbf{\textit{interpreting}}} ability of LMMs fake image by constructing \textbf{FakeClue}, which focuses on model's decision-making process and supports authenticity judgments in human-understandable terms respectively~\cite{zhao2024explainability}. Regarding the reasoning process, the models need to comprehend image contents and invoke internal knowledge to spot telltale clues to draw the final authenticity judgment, requiring a sophisticated multi-hop inference to analyze and correlate diverse evidence meticulously. Meanwhile, interpreting is the ability of models to articulate clear and understandable descriptions to support the given judgment, which facilitates greater transparency and trust in detection results. Henceforth, we provide two prompting modes in FakeClue, including \textit{fault-finding mode} and \textit{inference mode}. As depicted in Fig.~\ref{fig:overview}(b), the \textit{fault-finding mode} utilizes few-shot in-context prompting~\cite{rubin2021learning} by providing true image authenticity directly in the prompt, \emph{i.e.}, describing the visual forgery cues when queried by ``\textit{This is a fake image, explain why.}'' On the other hand, the \textit{inference mode} (Fig.~\ref{fig:overview}(c)) employs zero-shot chain-of-thought (CoT) prompting~\cite{WeiNIPS2022} that claims for explicit reasoning processes to analyze and correlate diverse evidence for the final authenticity conclusion. 

Collecting authenticity-related descriptions is arduous for both humans and machines since it requires visual insights and knowledge reserve. LMMs normally excel in extensive internal knowledge, while humans are good at logical reasoning. Therefore, we incorporate human-supervised LMM assistance under the guidance of the taxonomy for visual forgery proposed in Sec.~\ref{sec:forgerytaxonomy} to combine the strengths of both, which can unburden human workload and assure data reliability simultaneously. Specifically, as shown in Fig.~\ref{fig:goldenclues}, we adopt the human-in-the-loop strategy~\cite{wu2022survey} by initially prompting the proprietary GPT-4V~\cite{achiam2023gpt} in the \textit{fault-finding mode} with extra manual step-by-step demonstrations to produce primitive descriptions of fake clues for further refining. Then, we recruit human experts to verify and supplement the descriptions forthwith. Each expert is asked to judge the authenticity first and can only refine the telltale clues for the correctly recognized images because perception and understanding are no longer reliable if an image successively fools the human expert. In particular, human refinement is based on the following principles: \textbf{1)} Adding absent forgery descriptions that are obvious to human perception. \textbf{2)} Removing conflicting annotations against human perception. \textbf{3)} Reducing non-robust descriptions that do not uniquely exist in fake images. In this way, the remarkable logical inference ability of humans, together with the decent low-level vision and abundant internal knowledge of LMMs, can be centralized to make up the \textbf{golden clues} for fake images. 

As shown in Fig.~\ref{fig:goldenclues}, the golden clues in FakeClue, denoted by $\mathtt{C}$, serve as the demonstration of a coherent flow of sentences revealing the premises of the conclusion mimicking the human logical inference process on image authenticity. Each fake image is assigned at least four pieces of descriptions on forgery clues. The 6,000 $\mathtt{(I_0,\{Q\}_{i},C)}$ tuples constitute FakeClue, where $i\in\{1,2\}$ denotes \textit{fault-finding mode} and \textit{inference mode} respectively. Note that FakeClue contains no explanations for genuine images because the reasoning for fake images is to encompass and extend the recognition of real images. More examples of FakeClue data are provided in the supplementary materials.

\noindent{\textit{\textbf{Evaluation}}}\quad
As shown in Fig.~\ref{fig:overview}, the \textit{inference mode} is utilized to evaluate the reasoning ability of LMMs on FakeClue, where the generated explanations are compared against the golden clues and evaluated with automatic and GPT-assisted measures (see Sec.~\ref{sec:setups}). As for the interpretation ability where the standard authenticity is notified to models beforehand, model responses under the bare \textit{fault-finding mode} without manual guidance are evaluated for every fake image, quantified with the same measures.
\subsubsection{FakeQA}
\label{sec:fakeqa}
The \textbf{FakeQA} dataset tends to investigate the ability of LMMs to {\color{mygreen}\textbf{\textit{analyze}}} fine-grained forgery aspects. The data is in the $\mathtt{Q\&A}$ format on 14 dimensions of telltale clues (see Sec.~\ref{sec:forgerytaxonomy}), which analyzes image authenticity from fine-grained aspects. We utilize single-modal GPT to reorganize the golden clues into a unified $\mathtt{Q\&A}$ format, \textit{e.g.}, ``$\mathtt{Q}$: \textit{How is the texture regarding the authenticity of this image?} $\mathtt{A}$: \textit{The texture on the robot's surface is too uniform and lacks the variations you would expect from different materials.}''. These $\mathtt{Q\&A}$s maximally cover the spread of forgery aspects defined by the proposed taxonomy. In total, around 42k pieces of $\mathtt{(I_0,\{Q\}_{i},C)}$ tuples constitute FakeQA where $i=3$, indicating the \textit{free mode}.

\noindent{\textit{\textbf{Evaluation}}}\quad
We use the \textit{free mode} (Fig.~\ref{fig:overview}(d)) to probe LMMs on analyzing fine-grained forgery aspects. The evaluation is similar to that in FakeClue, where LMMs' responses are compared to the standard answers in FakeQA. Specifically, for each image, every query is tested within an individual ``\textless \textit{Image}\textgreater ~\textless \textit{Question}\textgreater'' context to minimize the mutual influence between questions. The measurements of the responses to all questions are averaged to indicate LMM's performance on each fake image, and the performance of all the fake images is averaged to reflect the performance on the entire FakeQA.
\begin{figure}[!tbp]
    \centering  
    \includegraphics[scale=0.14]{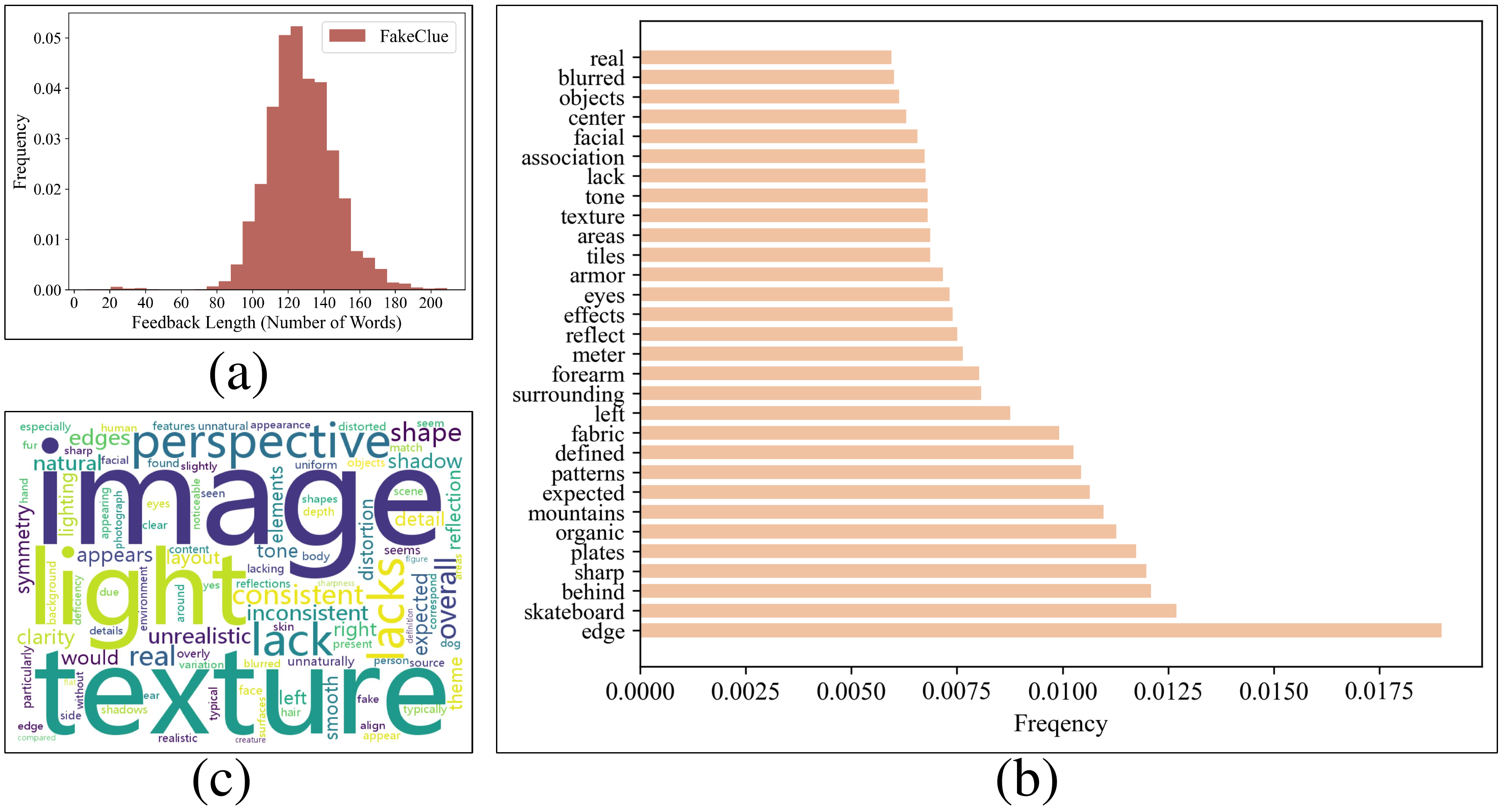}
    \caption{The illustration of word frequency statistics of FakeBench. (a) The distribution of feedback length in FakeClue. (b) Top-frequency national words of descriptions on forgery flaws in FakeClue. (c) Word-cloud of FakeClue.}
    \label{fig:statistics}
\end{figure}
\begin{table}[!tbp]
    \centering
    \fontsize{7}{8}\selectfont
    \caption{Key statistics of the \textbf{FakeBench}.}
    \begin{tabular}{lc}
    \toprule
    \makebox[0.25\textwidth][l]{Statistics} &Number\\
    \midrule
    Total Real/Fake Images &3,000/3,000 \\
    Total Questions &54,000 \\
    Generation Model Types &10 \\
    Images with Captions &6,000\\
    Difficulty Separation\\
    \quad \quad $\ast$ \textit{Easy: Medium: Hard} &1,241:969:790 \\
    \midrule
    Close-ended Questions &6,000\\
    \quad \quad $\ast$ \textit{Yes-or-No} Questions &3,000 \\
    \quad \quad $\ast$ \textit{What} Questions &3,000 \\
    Open-ended Questions &48,000 \\
    Questions with Long Explanations &6,000 \\
    \midrule
    Average Question Length &7 \\
    \quad \quad $\ast$ \textit{FakeClass} &9 \\
    \quad \quad $\ast$ \textit{FakeClue} &22 \\
    \quad \quad $\ast$ \textit{FakeQA} &10 \\
    Average Answer Length &49 \\
    \quad \quad $\ast$ \textit{FakeClass} &1 \\
    \quad \quad $\ast$ \textit{FakeClue} &127 \\
    \quad \quad $\ast$ \textit{FakeQA} &19 \\
    \bottomrule
    \end{tabular}
\label{tab:statistics}
\end{table}
\subsection{Statistical Analyses}
The detailed statistics of the proposed \textbf{FakeBench}, constituting \textbf{FakeClass}, \textbf{FakeClue}, and \textbf{FakeQA} datasets, are listed in Table~\ref{tab:statistics}. The answer length distribution and word frequency are illustrated in Fig.~\ref{fig:statistics}. Statistically, the average answer length in FakeClue reaches 127 words, enabling abundant descriptions of fake image flaws. Furthermore, we divide FakeBench into three difficulty levels according to the saliency level of image forgery based on human perception, \textit{i.e.}, the more salient the forgery in an image, the easier the sample. The detailed classification rules for difficulty levels are provided in the supplementary material. The distribution of \textit{Easy: Medium: Hard} samples is \textit{1241:969:790}, indicating a reasonable decrease in number as the difficulty increases. 

\section{Experimental Results}
Herein, we conduct extensive experiments and in-depth investigations on FakeBench. We introduce the employed baseline models and measures and then report experimental results and analyses on FakeBench. We further explore the effect of explicit reasoning guided by \textit{chain-of-thought prompting} on the accuracy of fake image detection. Last, several intriguing insights are presented. 
\subsection{Setups}
\label{sec:setups}
\begin{table}[!tbp]
    \centering
    \fontsize{7}{9}\selectfont
    \caption{Overview of the baseline open-source LMMs evaluated in the \textbf{FakeBench}. InternLM-XC.2-vl is short for InternLM-XComposer2-vl.}
    \setlength{\tabcolsep}{0.7mm}{\begin{tabular}{lccc}
    \toprule
    Model &Visual Model & V$\rightarrow$L Alignment & Language Model \\
    \midrule
    LLaVA-v1.5 (7B)~\cite{liu2024visual} & CLIP-ViT-Large/14 & MLP & Vicuna-7B\\
    Otter (7B)~\cite{li2023otter} & CLIP-ViT-Large/14 & Cross-Attention & MPT-7B \\
    mPLUG-Owl2 (7B)~\cite{ye2023mplugowl2} & CLIP-ViT-Large/14 & Visual Abstractor & LLaMA-7B\\
    Q-Instruct (7B)~\cite{wu2024q} &CLIP-ViT-Large/14 &Visual Abstractor &LLaMA-7B \\
    Qwen-VL (7B)~\cite{bai2023QwenVL} & Openclip-ViT-bigG & Cross-Attention & Qwen-7B \\
    Visual-GLM (6B)~\cite{du2022glm} & EVA-CLIP & Q-Former & ChatGLM-6B\\
    InstructBLIP (7B)~\cite{dai2023instructblip} & EVA-ViT-Large & Q-Former & Vicuna-7B\\
    IDEFICS-Instruct (7B)~\cite{laurençon2023idefics} & CLIP-ViT-Large/14 & Cross-Attention & LLaMA-7B\\
    Kosmos-2 (1B)~\cite{peng2023kosmos2} & CLIP-ViT-Large/14 & MLP & \textit{custom} (1B)\\
    InternLM-XC.2-vl (7B)~\cite{dong2024internlmxcomposer2} & CLIP-ViT-Large/14 &Partial LoRA  &InternLM2-7B\\
    \bottomrule
    \end{tabular}%
    }
\label{tab:LMM-baselines}
\end{table}
\subsubsection{Baselines} We evaluate 14 up-to-date LMMs on the entire FakeBench, including ten open-source models (\cite{liu2024visual,li2023otter,ye2023mplugowl2,bai2023QwenVL,du2022glm,dai2023instructblip,laurençon2023idefics,peng2023kosmos2,dong2024internlmxcomposer2,wu2024q}) and four closed-sourced ones (GPT-4V by OpenAI~\cite{achiam2023gpt}, GeminiPro by Google~\cite{geminiteam2024gemini}, Claude3 Sonnet and Claude3 Haiku by Antrophic~\cite{claude3}). The LMMs vary in the architecture and modality alignment strategy, detailed in Table~\ref{tab:LMM-baselines}. Besides, we further report the classification accuracy (ACC) of five unimodal models dedicated to fake image detection on FakeBench, including CNNSpot~\cite{wang2020cnn}, Gram-Net~\cite{liu2020global}, FreDect~\cite{frank2020leveraging}, PSM~\cite{ju2022fusing}, LGrad~\cite{tan2023learning}, and UnivDF~\cite{ojha2023towards}. To ensure fairness, all the LMMs and specialized image defaking models are evaluated under the \textbf{zero-shot setting} without further tuning on any additional data in FakeBench. Since the specialized models require fundamental training for classification, they are all trained under the generalization setting presented in \cite{wang2020cnn}, \textit{i.e.}, 360K fake images generated by ProGAN~\cite{karras2017progressive} and 360K real images from LSUN~\cite{yu2015lsun}. 
\subsubsection{Measures} Generally, LMM's accuracy in responding to the questions in FakeBench is utilized to signify their corresponding capabilities. We design rule-based systematic evaluation pipelines for both open- and closed-ended questions, which are elaborated in detail below. 

\noindent{\textbf{\textit{Closed-Ended}}}\quad 
For FakeClass, which is composed of closed-ended questions, we compute the correctness of the model's responses in comparison to the standard responses. The accuracy (ACC) is utilized as the metric to assess LMMs in fake image detection, which is computed by the ratio of correct responses to the total number of questions in FakeClass. 

\noindent{\textbf{\textit{Open-Ended}}}\quad For FakeClue and FakeQA encompassing open-ended questions, we compute text similarities under the narrative setting between model responses and the golden clues inspired by recent NLP studies~\cite{chiang2023can,zheng2024judging}. Aligned with previous work~\cite{lu2022learn,zhang2023multimodal,fu2023chain,wu2023q,liu2024visual,huang2024aesbench,yu2023mm}, we report automatic metrics (BLEU (B.)-1/2~\cite{papineni2002bleu}, ROUGE (R.)-L~\cite{lin2004rouge}, Sentence Similarity (Sim.)~\cite{reimers2019sentence}), and the LLM-as-a-judge method~\cite{zheng2024judging} to evaluate model responses. In particular, the LLM-as-a-judge evaluation focuses on three dimensions: (1) \textit{Completeness} (Comp.): More information aligning with the golden clues is preferred. (2) \textit{Preciseness} (Prec.): Information that conflicts with the golden clues will be penalized. (3) \textit{Relevance} (Rele.): More information should be closely related to image authenticity. For each image, the GPT-assisted measures select one value from \{0,1,2\}, indicating the low to high levels. We employ the proprietary GPT-4 to act as the evaluation agent, and the detailed prompts for evaluation are provided in the supplementary. The GPT-assisted measurement for each response is averaged across five rounds of repeated rating to ensure reliability, which is computed by
\begin{equation}
    S_{C,P,R} = \frac{\sum_{i=1}^{5}S_{c,p,r}}{5},
\end{equation}
where $s_{c,p,r}$ and $S_{C,P,R}$ denote the completeness, precision, and relevance ratings of each round and the averaged rating, respectively, and $i$ denotes the index of rating rounds.
Finally, we adopt macro-averaged automatic and normalized GPT4-assisted measures as each model's overall evaluation (Avr.).

\subsection{Key Results and Discoveries}
\begin{table*}[!t]
\caption{Evaluation results on the FakeClass dataset in accuracy (\%) regarding fake image \textbf{Detection} ability. The first and second best performances of LMMs are highlighted in \textbf{bold} and \underline{underlined}, respectively. The first and second best performances of dedicated models are highlighted in \textbf{\textit{BOLD ITALIC}} and \underline{\textit{UNDERLINED ITALIC}}. LMMs and dedicated models are respectively ranked. $\ast$ indicates proprietary models. The ``generation models'' columns are the decomposition of the ``fake'' column, and both the averaged ``authenticity'' columns and ``question type'' columns reflect the ``overall'' column.}  
\label{tab:fakeclass-result}
\renewcommand{\arraystretch}{1.1}
\fontsize{7pt}{8pt}\selectfont %
\centering
\setlength{\tabcolsep}{0.5mm}{
\begin{tabular}{l|cc|cc|cccccccccc|c|c}
\toprule
\multirow{2}*{\textbf{Model}}  & \multicolumn{2}{c|}{\textbf{Authenticity}}   & \multicolumn{2}{c|}{\textbf{Question Type}}  & \multicolumn{10}{c|}{\textbf{Generation Model}}   & {\multirow{2}*{\textbf{Overall}}}& {\multirow{2}*{\textbf{Rank}}}\\
\cmidrule(lr){2-3}\cmidrule(lr){4-5}\cmidrule(lr){6-15}
&\emph{Fake} &\emph{Real} &\emph{What} &\emph{Yes-or-No} &\emph{proGAN} &\emph{styleGAN} &\emph{CogView} &\emph{FuseDream} &\emph{VQDM} &\emph{Glide} &\emph{SD} &\emph{DALL$\cdot$E2}&\emph{DALL$\cdot$E3}&\emph{MJ}&& \\
\midrule 
\emph{Random guess}  &42.27\% &58.03\% &51.80\% &48.50\% &40.33\% &36.00\% &51.67\% &33.67\% &38.00\% &44.67\% &32.33\% &47.67\% &37.67\%&60.67\% &50.15\% &/   \\
\arrayrulecolor{gray}
\midrule
\textit{Humans (Best)} &97.00\% &93.00\% &/&/ &100.00\%&100.00\%&100.00\%&100.00\%&100.00\%&100.00\%&100.00\%&100.00\%&100.00\%&93.00\%&87.50\% &/\\
\textit{Humans (Worst)} &43.00\% &13.00\% &/&/&50.00\%&0.00\%&46.15\%&77.78\%&33.33\%&16.67\%&41.67\%&7.14\%&14.29\%&0.00\%&55.00\%&/\\
\textit{Humans (Overall)} &76.91\% &72.12\% &/&/&95.47\% &58.82\%  &87.07\% &95.90\% &63.96\% &82.91\% &79.41\% &77.66\% &70.69\% &45.59\% &74.51\% &/\\
\midrule
\rowcolor{gray!15}
\multicolumn{17}{l}{\textit{\textbf{LMMs (Zero-shot):}}}\\[0.1ex]
$\ast$ GPT-4V~\cite{achiam2023gpt} &\underline{59.87\%} &96.20\% &\textbf{76.33\%} &\textbf{79.73\%} &\textbf{95.00\%} &\textbf{66.00\%} &\underline{66.67\%} &\textbf{81.33\%} &\underline{36.67\%} &\underline{49.00\%} &45.67\% &44.33\% &61.00\% &53.00\% &\textbf{78.03\%} &\textbf{1}\\
mPLUG-Owl2~\cite{ye2023mplugowl2} &49.60\% &93.97\% &\underline{71.87\%} &71.70\%&56.33\% &10.00\% &\textbf{72.00\%} &\underline{75.33\%} &19.33\% &32.00\% &\underline{56.00\%} &\underline{55.00\%} &\underline{64.00\%} &\underline{56.00\%} &\underline{71.78\%} &\underline{2}\\
$\ast$ GeminiPro~\cite{geminiteam2024gemini} &35.83\% &\underline{99.27\%} &61.27\% &\underline{73.83\%} &45.33\% &15.67\% &47.33\% &52.67\% &19.33\% &28.00\% &35.33\% &35.67\% &41.00\% &38.00\%&67.50\% &3\\
Q-Instruct~\cite{wu2024q} &37.80\% &89.66\% &57.80\% &69.67\% &53.33\% &3.33\% &52.33\% &60.67\% &19.33\% &57.33\% &32.00\% &31.67\% &37.67\% &30.33\% &63.73\% &4\\
InternLM-XC.2-vl~\cite{dong2024internlmxcomposer2} &32.17\% &92.33\%&60.40\% &64.10\% &54.00\% &12.67\% &43.33\% &49.00\% &19.00\% &20.33\% &29.33\% &29.67\% &34.67\% &29.67\% &62.25\%&5\\
InstructBLIP~\cite{dai2023instructblip} &\textbf{67.80\%} &47.67\% &65.23\% &50.23\% &\underline{80.67\%} &\underline{65.33\%} &66.00\% &71.33\% &\textbf{68.00\%} &\textbf{70.33\%} &\textbf{60.33\%} &\textbf{61.00\%} &\textbf{70.67\%} &\textbf{64.33\%} &57.73\% &6\\
LLaVA-v1.5~\cite{liu2024visual} &38.00\% &77.40\% &55.77\% &59.63\% &44.33\% &17.33\% &45.33\% &55.33\% &23.00\% &28.33\% &39.67\% &0.00\% &0.00\% &41.00\% &57.70\%&7\\
Qwen-VL~\cite{bai2023QwenVL} &28.57\% &84.27\% &51.17\% &61.17\% &45.67\% &5.67\% &35.00\% &37.67\% &8.33\% &14.33\% &31.33\% &21.00\% &46.67\% &40.00\% &56.42\% &8 \\
$\ast$ Claude3 Sonnet~\cite{claude3} &12.00\% &98.23\% &55.97\% &54.27\% &7.00\% &0.00\% &20.33\% &13.67\% &0.67\% &2.67\% &15.33\% &9.33\% &30.33\% &20.67\% &55.12\% &9\\
$\ast$ Claude3 Haiku~\cite{claude3} &5.13\% &98.87\% &50.67\% &53.33\% &0.00\% &0.00\% &9.00\% &9.00\% &0.00\% &0.67\% &12.00\% &5.67\% &8.00\% &7.00\% &52.00\% &10\\
Otter~\cite{li2023otter} &0.40\% &\textbf{99.90\%} &50.03\% &50.27\% &0.33\% &0.00\% &0.00\% &0.67\% &0.67\% &0.33\% &0.33\% &0.00\% &0.67\% &1.00\% &50.15\% &11\\
Visual-GLM~\cite{du2022glm} &5.00\% &52.07\% &8.73\% &48.33\% &1.33\% &2.67\% &7.00\% &2.67\% &2.33\% &5.33\% &7.67\% &4.33\% &8.33\% &8.33\% &28.53\% &12\\
IDEFICS-Instruct~\cite{laurençon2023idefics} &24.97\% &31.97\% &7.00\% &49.93\% &19.67\% &27.33\% &24.33\% &28.33\% &27.67\% &24.00\% &23.67\% &23.67\% &24.00\% &27.00\% &28.47\% &13\\
Kosmos-2~\cite{peng2023kosmos2} &29.43\% &10.37\% &21.67\% &18.13\% &35.67\% &26.33\% &35.00\% &21.33\% &16.67\% &27.33\% &33.00\% &33.33\% &34.33\% &31.33\% &19.90\% &14\\
\arrayrulecolor{gray}
\midrule
\arrayrulecolor{black}
\rowcolor{gray!15}
\multicolumn{17}{l}{\textit{\textbf{DNN-based Specialized Models (Trained on CNNSpot~\cite{wang2020cnn} (ProGAN) with no overlapping of FakeBench):}}} \\
UnivDF~\cite{ojha2023towards} &32.35\% &\textbf{\textit{98.37\%}} &/&/&\underline{\textit{99.33\%}} &42.00\% &\textit{\underline{11.00\%}} &\textit{\underline{78.33\%}} &56.67\% &26.67\% &4.33\% &\textit{\underline{4.33\%}} &0.67\% &0.17\%&\textbf{\textit{65.36\%}} &\textbf{\textit{1}}\\
FreDect~\cite{frank2020leveraging} &\textbf{\textit{56.63\%}} &73.37\% &/&/&99.00\% &21.00\% &\textbf{54.33\%} &\textbf{\textit{82.00\%}} &\textbf{\textit{88.33\%}} &\textbf{\textit{41.67\%}} &\textbf{\textit{60.67\%}} &\textbf{\textit{58.67\%}} &\textbf{\textit{29.33\%}} &31.33\% &\textit{\underline{65.00\%}} &\textit{\underline{2}}\\
CNNSpot~\cite{wang2020cnn} &24.03\% &\textit{\underline{95.97\%}} &/&/&\textbf{\textit{100.00\%}} &\textbf{\textit{84.67\%}} &3.67\% &24.00\% &14.33\% &7.67\% &0.00\% &0.00\% &0.67\% &5.33\% &60.00\% &3\\
PSM~\cite{ju2022fusing} &18.93\% &94.43\% &/&/&\textbf{\textit{100.00\%}} &44.67\% &5.67\% &8.00\% &12.67\% &3.33\% &1.00\% &0.00\% &0.00\% &14.00\% &56.68\% &4\\
LGrad~\cite{tan2023learning} &36.87\% &66.47\% &/&/&\textbf{\textit{100.00\%}} &46.33\% &10.00\% &2.00\% &78.67\% &35.00\% &\textit{\underline{8.67\%}} &0.00\% &\textit{\underline{14.33\%}} &\textbf{\textit{73.67\%}} &51.67\% &5\\
Gram-Net~\cite{liu2020global} &\underline{\textit{36.93\%}} &65.50\%&/&/&\textbf{\textit{100.00\%}} &\textit{\underline{83.67\%}} &6.33\% &0.00\% &\textit{\underline{80.67\%}} &\textit{\underline{38.00\%}} &8.00\% &0.00\% &6.00\% &\textit{\underline{46.67\%}} &51.22\% &6\\
\bottomrule
\end{tabular}}
\end{table*}
\label{sec:results}
\subsubsection{Detection}
The {\color{mygreen}\textbf{\textit{detection}}} performance of 14 LMMs on FakeClass are shown in Table~\ref{tab:fakeclass-result}, where humans, LMMs, and specialized defaking models are compared. First, LMM's detection accuracy varies significantly, where GPT-4V and mPLUG-Owl2 achieve the best two overall performance, while Otter, Visual-GLM, IDEFICS-Instruct, and Kosmos-2 exhibit below-random performance. Second, most models perform better on the genuine image collection than on the fake, except for InstructBLIP and Kosmos-2. However, InstructBLIP achieves the best accuracy on fake images and multiple generator subsets simultaneously, exceeding specialized models. Claude3 Haiku, Sonnet, and Otter exhibit extreme discrepancy between the real and fake subsets, with notably high accuracy on the real subset and significantly lower accuracy on the fake subset, indicating they tend to classify any images as real ones. Third, most LMMs are robust to question formations, except for Visual-GLM and IDEFICS-Instruct, which prefer the \textit{Yes-or-No} questions more. Besides, the FakeBench is a great challenge for existing fake image detection models because of the great domain shifting concerning generators and image contents compared with previous datasets. 

Based on the observations, we obtain several intriguing insights for LMM's fake image {\color{mygreen}\textbf{\textit{detection}}} ability: \textbf{\textit{(a)}} The majority of the \textbf{LMMs can outperform \textit{random guess}} on the overall accuracy and even exceed the specialized models (\textit{e.g.}, GPT-4V, InstructBLIP, mPLUG-Owl2 and GeminiPro). Considering LMMs incorporate \textit{no explicit training} on deepfake detection, this implies a promising prospect for general-purpose large foundational models as more robust and generalizable fake image detectors with careful knowledge augmentation. \textbf{\textit{(b)}} Compared with open-source LMMs, \textbf{proprietary models exhibit slight superiority on detection}. However, even the strongest GPT-4V is outdone by the \textit{vanilla} InstructBLIP on the fake image set ($59.87\%<67.80\%$). This is an unexpected finding since commercial LMMs are widely acknowledged to have leading performances on various occasions. Fortunately, this indicates that model scale is not the only answer for the image defaking task, where open-source LMMs have promising potential to achieve promising performance when fine-tuned properly. \textbf{\textit{(c)}} \textbf{Humans far exceed models on image defaking.} Current Large Multimodal Models (LMMs) and specialized deep models lag behind humans in accuracy and generalization. Only the GPT-4V slightly outperforms the average human intelligence ($78.03\%>74.51\%$). However, all tested models significantly underperform compared to human peak performance, with the bottom five LMMs and two specialized models falling below the human minimum. This underscores a systematic advantage for humans in image defaking. \textbf{\textit{(d)}} Distinguishing the fake is a \textbf{harder} task for most LMMs than recognizing real ones. They excel with specific image generators but struggle with certain types, sometimes achieving zero accuracy. This deficiency in image defaking tasks stems from the universal oversight in training and fine-tuning LMMs, where the distinction between synthetic and genuine data is often neglected.

\subsubsection{Interpreting and Reasoning} LMM's performance on fake image {\color{mygreen}\textbf{\textit{interpreting}}} (effect-to-cause) and {\color{mygreen}\textbf{\textit{reasoning}}} (cause-to-effect) is exhibited in Table~\ref{tab:fakeclue-interpresult} and \ref{tab:fakeclue-reasonresult} respectively. These results provide an overall reflection of causal investigations of LMMs. We observe that GPT-4V and InstructBLIP achieve the top two performances among all models. Nonetheless, the majority of LMMs still struggle to reason and interpret. Notably, the LMMs tend to gather in lower score segments on the reasoning task, whereas the other 12 out of 14 models achieve the overall score in the $[0.357,0)$ range. Besides, all the models perform better in interpreting fake images than reasoning. In particular, LMMs tend to exhibit higher relevance scores on the interpretation task, indicating the potential to generate authenticity-relevant descriptions when correctly guided by the given image authenticity.

We draw several findings on the explainability-related abilities of LMMs from the observations: \textbf{\textit{(a)}} \textbf{Reasoning is far more difficult than interpreting} for LMMs on fake images. Few-shot prompting marginally improves forgery description by providing authenticity. However, constructing a coherent reasoning chain from these forgery clues remains a substantial challenge for current LMMs. \textbf{\textit{(b)}} \textbf{Open-source models} exhibit competitive performance compared to the
proprietary ones. For instance, InstructBLIP slightly outperforms GPT-4V in reasoning ($0.464>0.460$). This minor edge highlights the potential of further optimized LMMs as image defaking explainers with adequate fine-tuning data. \textbf{\textit{(c)}} LMMs generally exhibit \textbf{notable shortcomings on fake image reasoning and interpreting}. This is largely because they are trained predominantly on general-purpose visual-text data, with a notable lack of authenticity-focused data.
\begin{table}[!t]
\caption{Evaluation results on the fake image \textbf{interpretation} ability. The first and second best performances are highlighted in bold and underlined, respectively. $\ast$ indicates proprietary models.}  
\label{tab:fakeclue-interpresult}
\renewcommand{\arraystretch}{1.1}
\fontsize{6pt}{8pt}\selectfont %
\centering
\setlength{\tabcolsep}{0.6mm}{
\begin{tabular}{l|ccccc|cccc|c|c}
\toprule
\multirow{2}*{LMM}  & \multicolumn{5}{c|}{\textbf{Automatic Metrics}}  & \multicolumn{4}{c|}{\textbf{GPT-assisted Evaluation}}   & \multirow{2}*{\textbf{Avr.}} & {\multirow{2}*{\textbf{Rank}}}\\
\cmidrule(lr){2-6}\cmidrule(lr){7-10}
&\emph{B.-1} &\emph{B.-2} &\emph{R.-L} &\emph{Sim.} &\emph{Avr.}  &\emph{Comp.}&\emph{Prec.} &\emph{Rele.}  &\emph{Avr.} &&\\
\midrule 
$\ast$ GPT-4V~\cite{achiam2023gpt}  &\underline{0.188} &\textbf{0.092} &\textbf{0.211}&\textbf{0.635}&\textbf{0.282}&1.872&\textbf{1.423}&\underline{1.891}&1.729&0\textbf{.573}&\textbf{1}\\
InstructBLIP~\cite{dai2023instructblip} &\textbf{0.207}&\underline{0.072}&\underline{0.206}&0.463&\underline{0.237}&1.841&\underline{1.398}&1.829&1.689&\underline{0.541}&\underline{2}\\
$\ast$ Claude3 Haiku~\cite{claude3} &0.156 &0.059 &0.180 &\underline{0.489} &0.221&1.755&1.235&\textbf{1.916}&1.635&0.519&3\\
IDEFICS-Instruct~\cite{laurençon2023idefics} &0.093&0.032&0.140&0.367&0.158&\textbf{1.936}&1.344&1.789&1.690&0.501&4\\
$\ast$ Claude3 Sonnet~\cite{claude3} &0.130&0.042&0.170&0.467&0.202&1.516&1.182&1.884&1.527&0.483&5\\
InternLM-XC.2-vl~\cite{dong2024internlmxcomposer2} &0.095&0.032&0.142&0.402&0.168&\underline{1.899}&1.260&1.534&1.564&0.475&6\\
LLaVA-v1.5~\cite{liu2024visual} &0.143&0.047&0.159&0.466&0.204&1.812&0.964&1.621&1.466&0.468&7\\
Q-Instruct~\cite{wu2024q} &0.121 &0.045 &0.155 &0.477 &0.200&1.694&1.003&1.661&1.453&0.463&8\\
Qwen-VL~\cite{bai2023QwenVL} &0.056&0.018&0.096&0.322&0.116&1.853&1.149&1.485&1.496&0.432&9\\
mPLUG-Owl2~\cite{ye2023mplugowl2} &0.099&0.032&0.146&0.455&0.183&1.721&0.839&1.415&1.325&0.423&10\\
Visual-GLM~\cite{du2022glm} &0.157&0.048&0.177&0.447&0.207&1.533&0.851&1.389&1.258&0.418&11\\
$\ast$ GeminiPro~\cite{geminiteam2024gemini} &0.068 &0.029 &0.113 &0.459&0.167&1.688&1.024&1.271&1.328&0.415&12\\
Kosmos-2~\cite{peng2023kosmos2} &0.078&0.023&0.133&0.431&0.166&1.602&0.777&1.267&1.215&0.387&13\\
Otter~\cite{li2023otter} &0.032&0.007&0.067&0.287&0.098&0.952&1.456&1.492&1.300&0.374&14\\

\bottomrule
\end{tabular}}
\end{table}
\begin{table}[!t]
\caption{Evaluation results on the fake image \textbf{reasoning} ability. The first and second best performances are highlighted in bold and underlined, respectively. $\ast$ indicates proprietary models.}  
\label{tab:fakeclue-reasonresult}
\renewcommand{\arraystretch}{1.1}
\fontsize{6pt}{8pt}\selectfont %
\centering
\setlength{\tabcolsep}{0.5mm}{
\begin{tabular}{l|ccccc|cccc|c|c}
\toprule
\multirow{2}*{LMM}  & \multicolumn{5}{c|}{\textbf{Automatic Metrics}}  & \multicolumn{4}{c|}{\textbf{GPT-assisted Evaluation}}   & \multirow{2}*{\textbf{Avr.}} & {\multirow{2}*{\textbf{Rank}}}\\
\cmidrule(lr){2-6}\cmidrule(lr){7-10}
&\emph{B.-1} &\emph{B.-2} &\emph{R.-L} &\emph{Sim.} &\emph{Avr.}  &\emph{Comp.} &\emph{Prec.}  &\emph{Rele.} &\emph{Avr.} &&\\
\midrule
InstructBLIP~\cite{dai2023instructblip} &0.166&\underline{0.065}&0.160&0.458&0.212&\textbf{1.828}&\underline{0.922}&\textbf{1.546}&\textbf{1.432}&\textbf{0.464}&\textbf{1}\\
$\ast$ GPT-4V~\cite{achiam2023gpt} &\textbf{0.273} &\textbf{0.118} &\textbf{0.225} &\textbf{0.612} &\textbf{0.307}&1.392&0.777&\underline{1.503}&\underline{1.224}&\underline{0.460}&\underline{2}\\
$\ast$ Claude3 Sonnet~\cite{claude3} &0.188&0.062&\underline{0.193}&\underline{0.528}&\underline{0.243}&0.928&0.441&1.456&0.942&0.357&3\\
Visual-GLM~\cite{du2022glm} &\underline{0.198}&0.064&0.181&0.476&0.230&\underline{1.401}&0.408&1.094&0.968&0.357&4\\
InternLM-XC.2-vl~\cite{dong2024internlmxcomposer2} &0.108&0.032&0.169&0.492&0.200&1.004&0.670&1.167&0.947&0.337&5\\
Q-Instruct~\cite{wu2024q} &0.162 &0.052 &0.158 &0.484&0.214&0.963 &0.469&1.420&0.951 &0.335&6\\
$\ast$ GeminiPro~\cite{geminiteam2024gemini} &0.168 &0.059 &0.181 &0.485 &0.223&0.833&0.561&1.239&0.878&0.331&7\\
Kosmos-2~\cite{peng2023kosmos2} &0.115&0.045&0.112&0.319&0.148&0.972&0.900&1.176&1.016&0.328&8\\
$\ast$ Claude3 Haiku~\cite{claude3} &0.129 &0.041 &0.177 &0.512 &0.215&0.872&0.498&1.150&0.840&0.318&9\\
Otter~\cite{li2023otter}&0.030&0.007&0.060&0.266&0.091&0.903&\textbf{1.080}&1.251&1.078&0.315&10\\
LLaVA-v1.5~\cite{liu2024visual} &0.151&0.051&0.167&0.465&0.209&0.905&0.404&1.166&0.825&0.311&11\\
IDEFICS-Instruct~\cite{laurençon2023idefics} &0.099&0.026&0.148&0.457&0.183&1.178&0.450&0.967&0.865&0.308&12\\
Qwen-VL~\cite{bai2023QwenVL} &0.130&0.045&0.159&0.469&0.201&0.929&0.401&1.104&0.811&0.303&13\\
mPLUG-Owl2~\cite{ye2023mplugowl2} 
&0.109&0.034&0.160&0.458&0.190&0.843&0.500&1.070&0.804&0.296&14\\
\bottomrule
\end{tabular}}
\end{table}
\begin{table}[!tbp]
\caption{Evaluation results on the \textbf{fine-grained forgery analysis} ability. The first and second best performances are highlighted in bold and underlined, respectively. $\ast$ indicates proprietary models.}  
\label{tab:fakeqa-result}
\renewcommand{\arraystretch}{1.2}
\fontsize{6pt}{8pt}\selectfont %
\centering
\setlength{\tabcolsep}{0.5mm}{
\begin{tabular}{l|ccccc|cccc|c|c}
\toprule
\multirow{2}*{LMM}  & \multicolumn{5}{c|}{\textbf{Automatic Metrics}}  & \multicolumn{4}{c|}{\textbf{GPT-assisted Evaluation}}   & \multirow{2}*{\textbf{Avr.}} & {\multirow{2}*{\textbf{Rank}}}\\
\cmidrule(lr){2-6}\cmidrule(lr){7-10}
&\emph{B.-1} &\emph{B.-2} &\emph{R.-L} &\emph{Sim.} &\emph{Avr.} &\emph{Comp.} &\emph{Prec.} &\emph{Rele.}   &\emph{Avr.} &&\\
\midrule
$\ast$ GPT-4V~\cite{achiam2023gpt} &0.062 &0.022&0.177&0.455&0.179&\textbf{1.518}&\textbf{1.261}&1.869&\textbf{1.549}&\textbf{0.477}&\textbf{1}\\
$\ast$ Claude3 Sonnet~\cite{claude3} &0.038 &0.012 &0.158&0.398&0.152&1.254&1.091&\textbf{1.886}&1.410&0.429&\underline{2}\\
$\ast$ GeminiPro~\cite{geminiteam2024gemini} &0.178 &0.071&0.199&0.365&0.203&1.247&0.675&1.686&1.203&0.402&3\\
$\ast$ Claude3 Haiku~\cite{claude3} &0.097&0.045&0.202&\textbf{0.462}&0.202&1.071&0.734&1.771&1.192&0.399&4\\
LLaVA-v1.5~\cite{liu2024visual} &0.128&0.051&0.161&0.360&0.175&1.259 &0.757 &1.659&1.225&0.394&5\\
Q-Instruct~\cite{wu2024q} &0.150 &0.077&0.220 &0.394&0.210&1.240 &0.595 &1.573&1.136 &0.389&6\\
InternLM-XC.2-vl~\cite{dong2024internlmxcomposer2} &0.146&0.054&0.173&0.366&0.185&1.208&0.719&1.623&1.183&0.388&7\\
IDEFICS-Instruct~\cite{laurençon2023idefics} &0.189&0.086&0.231&0.381&0.222&1.165&0.456&1.483&1.035&0.370&8\\
mPLUG-Owl2~\cite{ye2023mplugowl2} &0.187&0.091&0.228&0.439&0.236&1.089 &0.412 &1.504 &1.002&0.368 &9\\
Otter~\cite{li2023otter} &0.144&0.055&0.179&0.412&0.198&1.088 &0.572 &1.519&1.060&0.364&10\\
Kosmos-2~\cite{peng2023kosmos2} &\textbf{0.216}&\textbf{0.104}&\textbf{0.251}&0.458&\textbf{0.257}&1.028&0.371&1.342&0.914&0.357&11\\
Visual-GLM~\cite{du2022glm} &0.029&0.011&0.142&0.394&0.144&1.085&0.603&1.228&0.972&0.315&12\\
Qwen-VL~\cite{bai2023QwenVL} &0.076&0.026&0.105&0.241&0.112&1.004&0.490&0.802&0.765&0.247&13\\
InstructBLIP~\cite{dai2023instructblip} &0.004&0.001&0.007&0.079&0.023&0.810&0.321&0.543&0.558&0.151&14\\
\bottomrule
\end{tabular}}
\end{table}
\subsubsection{Fine-Grained Forgery Analyses}FakeQA probes LMMs on analyzing fine-grained generative forgery clues of fake images, and the performances of LMMs are listed in Table~\ref{tab:fakeqa-result}. From the results of {\color{mygreen}\textbf{\textit{fine-grained forgery analyses}}}, we observe that GPT-4V again outperforms others. Moreover, the propriety LMMs, including GPT-4V, Claude3 Sonnet, and Claude3 Haiku, all exhibit superiority over their open-source counterparts. Besides, unlike the challenging reasoning task, InstructBLIP, Qwen-VL, and Visual-GLM exhibit even worse performance on FakeQA. In practice, they usually offer empty responses when required to analyze the forgery of a certain aspect, which can be reflected by their notably low completeness, BLUE, and ROUGE-L scores. 

According to the observations, we discover that: \textbf{\textit{(a)}} \textbf{Proprietary LMMs} typically outperform open-source ones in analyzing specific generative forgery aspects, though their best performance is still suboptimal (highest score of 0.477 out of 1). This reflects a fundamental deficiency in how current LMMs perceive and understand authenticity. \textbf{\textit{(b)}} For some LMMs, analyzing fine-grained forgery is more \textbf{challenging} than broad reasoning. This difficulty arises because current LMMs lack the fine-grained conceptual framework for in-depth authenticity analysis. \textbf{\textit{(c)}} Overall, most LMMs still struggle in fine-grained forgery analysis, highlighting a \textbf{systematic inadequacy} concerning concrete forensics knowledge in current training data used for LMMs.
\subsubsection{Inter-task Relevance}
We witness performance discrepancies from Table~\ref{tab:fakeclass-result} to \ref{tab:fakeqa-result}, highlighting the varied capabilities across defaking-related subtasks. The highest Spearman Rank Correlation (SRCC) among model rankings across four criteria is 0.54, noted between reasoning and fine-grained forgery analysis, suggesting these tasks share operational mechanisms. In contrast, the correlation between interpreting and reasoning is notably low at 0.046, underscoring these tasks' vastly different cognitive demands. This discrepancy emphasizes the specialized strengths and limitations of LMMs in the tasks requiring nuanced \textit{understanding and logical deduction} versus those demanding broad contextual \textit{interpreting}. 
\begin{table*}[!t]
\caption{Evaluation results on the influence of explicit reasoning on the performance of fake image detection in terms of accuracy, precision, and recall scores. The \emph{w} and \emph{w/o} respectively denote the authenticity judgment obtained \textit{with} or \textit{without} explicit reasoning ahead. The results with improvement are highlighted in \textbf{bold}.}  
\label{tab:reasoning-effect}
\renewcommand{\arraystretch}{1.1}
\fontsize{6.5pt}{8pt}\selectfont
\centering
\setlength{\tabcolsep}{1.3mm}{
\begin{tabular}{l|cc|cc|cc|cc|cc}
\toprule
\multirow{2}*{LMM}  & \multicolumn{2}{c|}{\textbf{Fake Image ACC}$\uparrow$}  & \multicolumn{2}{c|}{\textbf{Real Image ACC}$\uparrow$}   & \multicolumn{2}{c|}{\textbf{Overall ACC}$\uparrow$} &\multicolumn{2}{c|}{\textbf{Overall Precision}$\uparrow$}&\multicolumn{2}{c}{\textbf{Overall Recall}$\uparrow$}\\
\cmidrule(lr){2-3}\cmidrule(lr){4-5}\cmidrule(lr){6-7}\cmidrule(lr){8-9}\cmidrule(lr){10-11}
&\emph{w/o} &\emph{w} & \emph{w/o} &\emph{w} &\emph{w/o} &\emph{w} &\emph{w/o} &\emph{w} &\emph{w/o} &\emph{w}\\
\midrule 
\rowcolor{gray!15}
\multicolumn{11}{l}{\textit{\textbf{Closed-sourced LMMs:}}}\\[0.1ex]
GPT-4V~\cite{achiam2023gpt} & 59.87\% & 53.70\% (-6.17\%) & 96.20\% & 95.80\% (-0.40\%) & 78.03\% & 74.71\% (-3.32\%) & 97.66\% & 94.98\% (-2.68\%) & 60.01\% & 55.79\% (-4.22\%)\\
GeminiPro~\cite{geminiteam2024gemini} & 35.83\% & 27.30\% (-8.53\%) & 99.27\% & 96.37\% (-2.90\%) & 67.50\% & 61.82\% (-5.68\%) & 99.08\% & 98.20\% (-0.88\%) & 35.87\% & 28.15\% (-7.72\%)\\
Claude3 Sonnet~\cite{claude3} & 12.00\% & \textbf{13.43\% (+1.40\%)} & 98.23\% & \textbf{98.50\% (+0.27\%)} & 55.12\% & \textbf{55.95\% (+0.83\%)} & 92.31\% & \textbf{94.16\% (+1.85\%)} & 12.02\% & \textbf{13.52\% (+1.50\%)}\\
Claude3 Haiku~\cite{claude3} & 5.13\% & 2.20\% (-3.93\%) & 98.87\% & \textbf{99.23\% (+0.36\%)} & 52.00\% & 50.71\% (-1.29\%) & 89.53\% & \textbf{90.41\% (+0.88\%)} & 5.13\% & 2.20\% (-2.93\%)\\
\arrayrulecolor{gray}
\midrule 
\rowcolor{gray!15}
\multicolumn{11}{l}{\textit{\textbf{Open-source LMMs:}}}\\[0.1ex]
\arrayrulecolor{black}
mPLUG-Owl2~\cite{ye2023mplugowl2} & 49.60\% & 24.33\% (-25.27\%) & 93.97\% & \textbf{97.03\% (+3.06\%)} & 71.78\% & 60.68\% (-11.10\%) & 89.21\% & 89.13\% (-0.08\%) & 49.62\% & 24.33\% (-25.29\%)\\
Q-Instruct~\cite{wu2024q} &37.80\% &3.23\% (-34.57\%) &89.66\% &\textbf{99.43\% (+9.77\%)} &63.73\% &51.33\% (-12.40\%) &78.59\% &\textbf{85.09\% (+6.50\%)} &37.80\% &9.07\% (-28.73\%)\\
InternLM-XC.2-vl~\cite{dong2024internlmxcomposer2} & 32.17\% & 6.10\% (-26.07\%) & 92.33\% & \textbf{96.60\% (+4.27\%)} & 62.25\% & 51.34\% (-10.91\%) & 80.82\% & \textbf{93.85\% (+13.03\%)} & 32.19\% & 6.51\% (-25.68\%)\\
InstructBLIP~\cite{dai2023instructblip} & 67.80\% & 4.67\% (-63.13\%) & 47.67\% & 0.90\% (-46.77\%) & 57.73\% & 2.78\% (-54.95\%) & 56.45\% & 45.75\% (-10.70\%) & 67.85\% & \textbf{86.42\% (+18.57\%)}\\
LLaVA-v1.5~\cite{liu2024visual} & 38.00\% & 29.27\% (-8.73\%) & 77.40\% & \textbf{83.43\% (+6.03\%)} & 57.70\% & 56.35\% (-1.35\%) & 64.48\% & \textbf{64.65\% (+0.17\%)} & 38.75\% & 29.52\% (-9.23\%)\\
Qwen-VL~\cite{bai2023QwenVL} & 28.57\% & 5.93\% (-22.64\%) & 84.27\% & 20.71\% (-63.56\%) & 56.42\% & 13.32\% (-43.10\%) & 64.53\% & \textbf{91.28\% (+26.75\%)} & 28.62\% & 14.72\% (-13.90\%)\\
Otter~\cite{li2023otter} & 0.40\% & \textbf{40.88\% (+40.48\%)} & 99.90\% & 59.92\% (-39.98\%) & 50.15\% & \textbf{50.40\% (+0.25\%)} & 100.00\% & 57.18\% (-42.82\%) & 0.40\% & \textbf{45.16\% (+44.76\%)}\\
Visual-GLM~\cite{du2022glm} & 5.00\% & \textbf{9.67\% (+4.67\%)} & 52.07\% & 2.60\% (-49.47\%) & 28.53\% & 6.13\% (-22.40\%) & 62.50\% & \textbf{63.88\% (+1.38\%)} & 10.05\% & \textbf{83.57\% (+73.52\%)}\\
IDEFICS-Instruct~\cite{laurençon2023idefics} & 24.97\% & 14.67\% (-10.30\%) & 31.97\% & \textbf{79.57\% (+47.60\%)} & 28.47\% & \textbf{47.12\% (+18.65\%)} & 49.97\% & \textbf{92.44\% (+42.47\%)} & 46.78\% & 16.55\% (-30.23\%)\\
Kosmos-2~\cite{peng2023kosmos2} & 29.43\% & 6.97\% (-22.46\%) & 10.37\% & 4.10\% (-6.27\%) & 19.90\% & 5.54\% (-14.36\%) & 54.17\% & 49.53\% (-4.64\%) & 63.03\% & \textbf{81.64\% (+18.61\%)}\\
\bottomrule
\end{tabular}}
\end{table*}
\subsection{Influence of Fine-Tuning Data} 
Q-Instruct~\cite{wu2024q} derived from mPLUG-Owl2, focuses on enhancing the low-level vision capabilities of foundation LMMs, such as image quality assessment and distortion tracing. According to Table~\ref{tab:fakeclass-result} to \ref{tab:fakeqa-result}, Q-Instruct demonstrates significantly improved performance on FakeClue and FakeQA compared to its predecessor, mPLUG-Owl2, illustrating the benefits of instruction tuning. However, it also experiences a noticeable 8.05\% overall drop on \textbf{FakeClass}, particularly with fake images. This suggests that while enhancing LMMs with low-level vision knowledge improves their explanatory ability regarding image authenticity, it may compromise their accuracy in fake detection.
\subsection{Influence of Explicit Reasoning on Forgery Clues}
\begin{figure*}[!t]
    \centering  
    \includegraphics[scale=0.26]{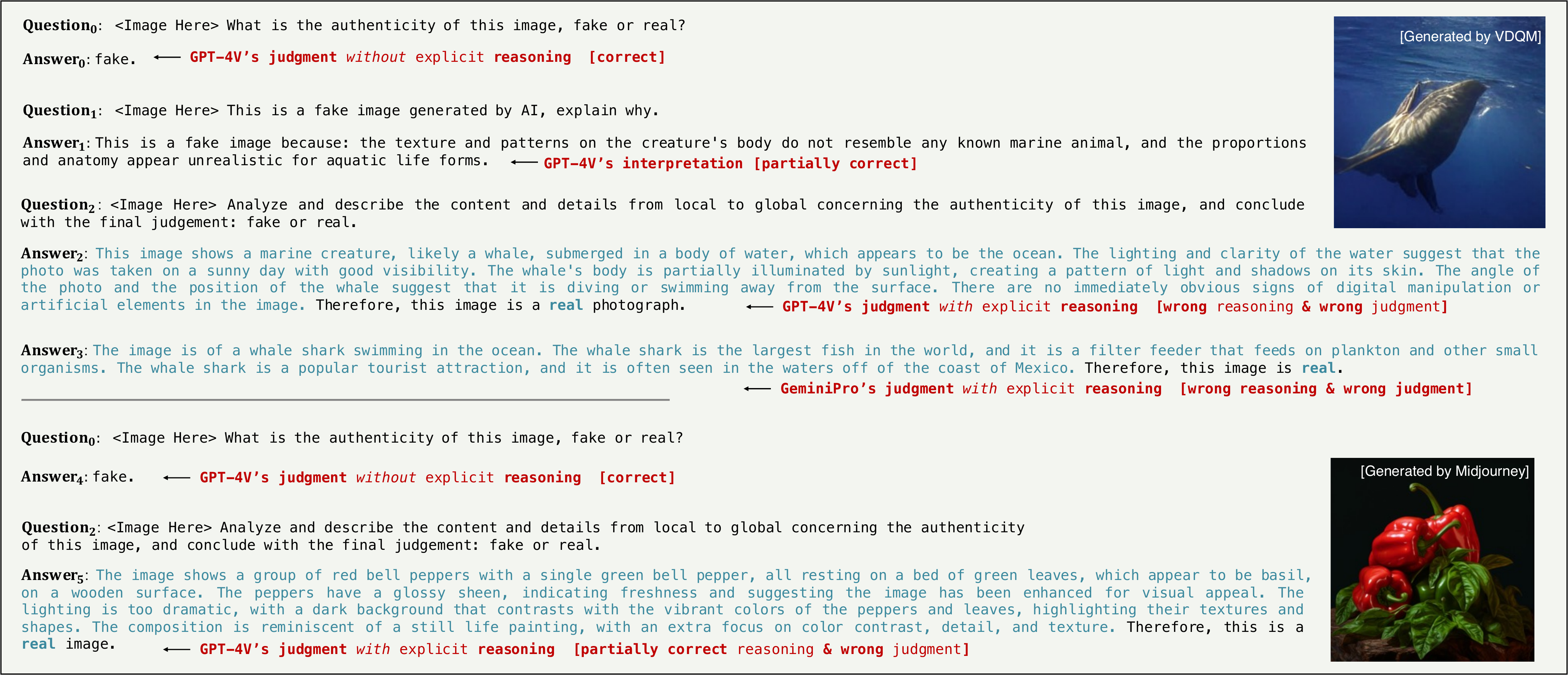}
    \caption{Comparisons of the responses from GPT-4V and GeminiPro in the \textit{basic mode} for detection, \textit{faultfinding mode} for interpretation and \textit{inference mode} for reasoning respectively.}
    \label{fig:casestudy}
\end{figure*}
In this section, we delve deeper into how explicit reasoning on forgery clues affects LMMs' performance in image defaking by comparing results from \textbf{FakeClass} \textit{with} and \textit{without} CoT prompting. Specifically, we calculate the ACC of detection under the \textit{inference mode} (see Sec.~\ref{sec:fakeclue}) that would ask for an authenticity judgment after the chain-of-thought reasoning and compare it with the accuracy under the \textit{basic mode} without CoT (see Sec.~\ref{sec:fakeclass}). Nonetheless, LMMs occasionally produce \textit{irrelevant answers} that are neither fake nor real (\textit{e.g.}, ``It is very hard to decide.''). Therefore, to assess the impact more comprehensively, we also report the Precision and Recall metrics on the FakeClass dataset. We treat fake images as the positive class and genuine images as the negative class, with the following calculations:
\begin{equation}
    Precision = \frac{TP}{TP+FP},\ Recall = \frac{TP}{TP+FN},
\end{equation}
where $TP$, $FP$, and $FN$ denote true positives, false positives, and false negatives, respectively. A higher precision score indicates fewer real images wrongly classified as fake, and a higher recall score indicates fewer fake images wrongly classified as real. All the results are listed in Table~\ref{tab:reasoning-effect}.
\subsubsection{Effects}
According to Table~\ref{tab:reasoning-effect}, \textbf{explicit reasoning before judgment significantly impacts LMM's performance on detection}. However, the influence varies considerably among different models, exhibiting positive or negative effects. Specifically, 7 out of 14 models exhibit improved accuracy on genuine images, while only three models show improvements on fake images. Only Claude3 Sonnet attains improvements across all subsets and measures with CoT, albeit subtly. IDEFICS-Instruct and Claude3 Sonnet can see overall accuracy improvements with CoT prompting, where the former experienced a decrease in fake images but a substantial increase in genuine images. Conversely, Otter shows a 40.48\% increase in accuracy on the fake image set yet suffered a significant drop on genuine images. In stark contrast, the other 6 out of 14 models have adverse impacts to different extents, with decreases in real and fake images. In particular, Qwen\mbox{-}VL loses 63.56\% in accuracy on fake images. Regarding precision and recall, eight models showed improvements in precision and five in recall. The inconsistency among the metrics is due to the irrelevant responses from LMMs in some cases.

Therefore, \textbf{the influence of explicit reasoning is various and inconsistent and negatively affects LMM's performance on image defaking in most cases.} It sometimes induces models to become overly picky in detecting the fake and thus erroneously label images as real (\textit{e.g.}, Q-Instruct, InterLM-XC.2-vl, LLaVA-v1.5, Otter, Visual-GLM). Also, it can globally impair detection capabilities (\textit{e.g.}, GPT-4V, GeminiPro), or hinder LMMs from providing relevant responses (\textit{e.g.}, InstructBLIP, Kosmos-2, Qwen-VL). However, as shown by $Answer_{2,3,5}$ of Fig.~\ref{fig:casestudy}, CoT prompting can enrich authenticity assessments by fostering a broader and deeper level of visual comprehension, where additional semantics, contexts, and even external knowledge contribute to authenticity inferences. This suggests that LMMs can integrate various informational aspects beyond visual perception, potentially enhancing image authenticity judgments if properly directed through fine-tuning. Refer to the supplementary for exemplar cases of positive and negative effects from explicit reasoning.
\subsubsection{Causes}
The ineffectiveness of CoT prompting in our study is unexpected, as explicit reasoning generally benefits various tasks across different domains, \textit{e.g.}, causal inference~\cite{WeiNIPS2022}, commonsense reasoning~\cite{zhang2022automatic} and image quality assessment~\cite{wu2024comprehensive}. The reasons for this anomaly appear to vary across models and can be categorized into three main aspects: \textbf{\textit{(a)}} \textbf{Limited understanding of generative cues.} LMMs often demonstrate insufficient visual reasoning capabilities concerning authenticity, as evidenced by their struggle to generate valid descriptions related to authenticity, even when explicitly prompted by CoT. This issue is highlighted in $Answer_2$ of Fig.~\ref{fig:casestudy}, where essential reasoning processes fail to be effectively conducted, leading to flawed descriptions and, consequently, erroneous judgments due to error propagation. \textbf{\textit{(b)}} \textbf{Defective projection from descriptions to authenticity.} Some models, like GPT-4V, can identify aspects of forgery but fail to correlate these descriptions accurately with authenticity. For instance, as shown in $Answer_5$ of Fig.~\ref{fig:casestudy}, GPT-4V identifies an imaging abnormality but ultimately makes an incorrect judgment, resulting in decreased detection performance. \textbf{\textit{(c)}} \textbf{Limited instruction-following ability.} The intrinsic ability of some models to follow instructions is compromised by insufficient instruction tuning, which hampers their capacity to adhere accurately to input queries. \textbf{\textit{(d)}} \textbf{Reliance on detection shortcuts.} LMMs could have \textit{shortcuts} in fake image detection. Fake image detection for LMMs is fundamentally a classification problem under the open-world setting and thus might partly depend on human-invisible cues left by generators, which are commonly exploited by DNN-based models~\cite{chai2020makes, ojha2023towards}. 

In summary, CoT-guided explicit reasoning divides the initial problem into sub-problems to be solved sequentially. However, difficulties in resolving sub-problems can collectively hinder the overall solution.
Consequently, the observed decline in the effectiveness of CoT prompting in our research further reflects LMMs' \textbf{defective reasoning} capabilities in image authenticity assessment, encompassing both the recognition of generative forgery and the projection to authenticity. Moreover, LMMs might have alternative pathways, aside from reason-to-judge, for detecting fake images, which remains an open question in LLM research. Investigating whether their judgments are more based on invisible cues left by generators or more on visual inconsistencies at various semantic levels could further benefit their detection capability. 

\section{Conclusions}
In this paper, we have introduced \textbf{FakeBench}, a pioneering multimodal benchmark tailored for probing LMMs on explainable fake image detection. We comprehensively investigated LMMs' performance across three components-- \textbf{FakeClass}, \textbf{FakeClue}, and \textbf{FakeQA} datasets, which correspond to \textit{detection}, \textit{causal investigations}, and fine-grained forgery analysis respectively. The experiments demonstrate that some LMMs have begun showing capabilities in explainable image defaking, although their effectiveness is not consistently balanced across varied scenarios. Additionally, we discovered that model scale is not a limiting factor; open-source LMMs can perform on par with their proprietary counterparts in certain respects. This presents an opportunity to enhance LMMs as transparent fake image detectors through targeted fine-tuning and knowledge augmentation. However, the authenticity-related knowledge is found hard to derive from the data of general-purpose visual tasks, highlighting the demand for dedicated datasets on image authenticity. Furthermore, our exploration of the chain-of-thought prompting revealed a general ineffectiveness of explicit reasoning on forgery, attributed to deficient authenticity reasoning capabilities. This underscores the critical need to integrate defaking-specific knowledge into LMMs to establish them as reliable detectors and explainers for fake images. Regarding the opportunity for future work, the current prompting framework can be further enriched to explore LMMs in a wider spread of contextual scenarios. We hope our findings with FakeBench could inspire more robust, transparent, and interpretable fake image detection systems, thereby advancing image forensics and AI risk management.


\bibliographystyle{IEEEtran}
\bibliography{refs}

\begin{thebibliography}{100}
\providecommand{\url}[1]{#1}
\csname url@samestyle\endcsname
\providecommand{\newblock}{\relax}
\providecommand{\bibinfo}[2]{#2}
\providecommand{\BIBentrySTDinterwordspacing}{\spaceskip=0pt\relax}
\providecommand{\BIBentryALTinterwordstretchfactor}{4}
\providecommand{\BIBentryALTinterwordspacing}{\spaceskip=\fontdimen2\font plus
\BIBentryALTinterwordstretchfactor\fontdimen3\font minus \fontdimen4\font\relax}
\providecommand{\BIBforeignlanguage}[2]{{%
\expandafter\ifx\csname l@#1\endcsname\relax
\typeout{** WARNING: IEEEtran.bst: No hyphenation pattern has been}%
\typeout{** loaded for the language `#1'. Using the pattern for}%
\typeout{** the default language instead.}%
\else
\language=\csname l@#1\endcsname
\fi
#2}}
\providecommand{\BIBdecl}{\relax}
\BIBdecl

\bibitem{midjourney}
\BIBentryALTinterwordspacing
``Midjourney,'' 2024. [Online]. Available: \url{https://www.midjourney.com/home}
\BIBentrySTDinterwordspacing

\bibitem{betker2023improving}
J.~Betker, G.~Goh, L.~Jing, T.~Brooks, J.~Wang, L.~Li, L.~Ouyang, J.~Zhuang, J.~Lee, Y.~Guo \emph{et~al.}, ``Improving image generation with better captions,'' \emph{Computer Science}, vol.~2, no.~3, p.~8, 2023.

\bibitem{barni2020cnn}
M.~Barni, K.~Kallas, E.~Nowroozi, and B.~Tondi, ``{CNN} detection of {GAN}-generated face images based on cross-band co-occurrences analysis,'' in \emph{WIFS}.\hskip 1em plus 0.5em minus 0.4em\relax IEEE, 2020, pp. 1--6.

\bibitem{frank2020leveraging}
J.~Frank, T.~Eisenhofer, L.~Sch{\"o}nherr, A.~Fischer, D.~Kolossa, and T.~Holz, ``Leveraging frequency analysis for deep fake image recognition,'' in \emph{ICML}.\hskip 1em plus 0.5em minus 0.4em\relax PMLR, 2020, pp. 3247--3258.

\bibitem{gragnaniello2021gan}
D.~Gragnaniello, D.~Cozzolino, F.~Marra, G.~Poggi, and L.~Verdoliva, ``Are {GAN} generated images easy to detect? {A} critical analysis of the state-of-the-art,'' in \emph{ICME}.\hskip 1em plus 0.5em minus 0.4em\relax IEEE, 2021, pp. 1--6.

\bibitem{ju2022fusing}
Y.~Ju, S.~Jia, L.~Ke, H.~Xue, K.~Nagano, and S.~Lyu, ``Fusing global and local features for generalized {AI}-synthesized image detection,'' in \emph{ICIP}.\hskip 1em plus 0.5em minus 0.4em\relax IEEE, 2022, pp. 3465--3469.

\bibitem{liu2020global}
Z.~Liu, X.~Qi, and P.~H. Torr, ``Global texture enhancement for fake face detection in the wild,'' in \emph{CVPR}.\hskip 1em plus 0.5em minus 0.4em\relax IEEE, 2020, pp. 8060--8069.

\bibitem{liu2022detecting}
B.~Liu, F.~Yang, X.~Bi, B.~Xiao, W.~Li, and X.~Gao, ``Detecting generated images by real images,'' in \emph{ECCV}.\hskip 1em plus 0.5em minus 0.4em\relax Springer, 2022, pp. 95--110.

\bibitem{tan2023learning}
C.~Tan, Y.~Zhao, S.~Wei, G.~Gu, and Y.~Wei, ``Learning on gradients: {G}eneralized artifacts representation for gan-generated images detection,'' in \emph{CVPR}.\hskip 1em plus 0.5em minus 0.4em\relax IEEE, 2023, pp. 12\,105--12\,114.

\bibitem{tariang2024synthetic}
D.~Tariang, R.~Corvi, D.~Cozzolino, G.~Poggi, K.~Nagano, and L.~Verdoliva, ``Synthetic image verification in the era of generative artificial intelligence: What works and what isn’t there yet,'' \emph{IEEE Security \& Privacy}, 2024.

\bibitem{dong2022explaining}
S.~Dong, J.~Wang, J.~Liang, H.~Fan, and R.~Ji, ``Explaining deepfake detection by analysing image matching,'' in \emph{ECCV}.\hskip 1em plus 0.5em minus 0.4em\relax Springer, 2022, pp. 18--35.

\bibitem{AghasanliICCVW2023}
A.~Aghasanli, D.~Kangin, and P.~Angelov, ``Interpretable-through-prototypes deepfake detection for diffusion models,'' in \emph{ICCVW}.\hskip 1em plus 0.5em minus 0.4em\relax IEEE, 2023, pp. 467--474.

\bibitem{chai2020makes}
L.~Chai, D.~Bau, S.-N. Lim, and P.~Isola, ``What makes fake images detectable? {U}nderstanding properties that generalize,'' in \emph{ECCV}.\hskip 1em plus 0.5em minus 0.4em\relax Springer, 2020, pp. 103--120.

\bibitem{bird2024cifake}
J.~J. Bird and A.~Lotfi, ``Cifake: Image classification and explainable identification of ai-generated synthetic images,'' \emph{IEEE Access}, 2024.

\bibitem{zhang2023perceptual}
L.~Zhang, Z.~Xu, C.~Barnes, Y.~Zhou, Q.~Liu, H.~Zhang, S.~Amirghodsi, Z.~Lin, E.~Shechtman, and J.~Shi, ``Perceptual artifacts localization for image synthesis tasks,'' in \emph{CVPR}.\hskip 1em plus 0.5em minus 0.4em\relax IEEE, 2023, pp. 7579--7590.

\bibitem{zhang2024common}
Y.~Zhang, B.~Colman, A.~Shahriyari, and G.~Bharaj, ``Common sense reasoning for deep fake detection,'' in \emph{ECCV}.\hskip 1em plus 0.5em minus 0.4em\relax Springer, 2024, pp. 1--1.

\bibitem{sun2023towards}
K.~Sun, S.~Chen, T.~Yao, H.~Yang, X.~Sun, S.~Ding, and R.~Ji, ``Towards general visual-linguistic face forgery detection,'' \emph{arXiv preprint arXiv:2307.16545}, 2023.

\bibitem{wang2020cnn}
S.-Y. Wang, O.~Wang, R.~Zhang, A.~Owens, and A.~A. Efros, ``{CNN}-generated images are surprisingly easy to spot... for now,'' in \emph{CVPR}.\hskip 1em plus 0.5em minus 0.4em\relax IEEE, 2020, pp. 8695--8704.

\bibitem{WangICCV2023}
Z.~Wang, J.~Bao, W.~Zhou, W.~Wang, H.~Hu, H.~Chen, and H.~Li, ``Dire for diffusion-generated image detection,'' in \emph{ICCV}.\hskip 1em plus 0.5em minus 0.4em\relax IEEE, 2023, pp. 22\,388--22\,398.

\bibitem{ZhongArxiv2023}
N.~Zhong, Y.~Xu, Z.~Qian, and X.~Zhang, ``Rich and poor texture contrast: A simple yet effective approach for ai-generated image detection,'' \emph{arXiv preprint arXiv:2311.12397v2}, 2023.

\bibitem{hou2023evading}
Y.~Hou, Q.~Guo, Y.~Huang, X.~Xie, L.~Ma, and J.~Zhao, ``Evading deepfake detectors via adversarial statistical consistency,'' in \emph{CVPR}.\hskip 1em plus 0.5em minus 0.4em\relax IEEE, 2023, pp. 12\,271--12\,280.

\bibitem{ojha2023towards}
U.~Ojha, Y.~Li, and Y.~J. Lee, ``Towards universal fake image detectors that generalize across generative models,'' in \emph{CVPR}.\hskip 1em plus 0.5em minus 0.4em\relax IEEE, 2023, pp. 24\,480--24\,489.

\bibitem{xu2023exposing}
Q.~Xu, H.~Wang, L.~Meng, Z.~Mi, J.~Yuan, and H.~Yan, ``Exposing fake images generated by text-to-image diffusion models,'' \emph{Pattern Recogn. Lett.}, vol. 176, pp. 76--82, 2023.

\bibitem{liu2024visual}
H.~Liu, C.~Li, Q.~Wu, and Y.~J. Lee, ``Visual instruction tuning,'' in \emph{NeurIPS}, vol.~36, 2024.

\bibitem{bai2023QwenVL}
J.~Bai, S.~Bai, S.~Yang, S.~Wang, S.~Tan, P.~Wang, J.~Lin, C.~Zhou, and J.~Zhou, ``Qwen-{VL}: A versatile vision-language model for understanding, localization, text reading, and beyond,'' \emph{arXiv preprint arXiv:2308.12966}, 2023.

\bibitem{dai2023instructblip}
W.~Dai, J.~Li, D.~Li, A.~M.~H. Tiong, J.~Zhao, W.~Wang, B.~Li, P.~Fung, and S.~Hoi, ``Instructblip: Towards general-purpose vision-language models with instruction tuning,'' in \emph{NeurIPS}, 2023.

\bibitem{dosovitskiy2020image}
A.~Dosovitskiy, L.~Beyer, A.~Kolesnikov, D.~Weissenborn, X.~Zhai, T.~Unterthiner, M.~Dehghani, M.~Minderer, G.~Heigold, S.~Gelly \emph{et~al.}, ``An image is worth 16x16 words: Transformers for image recognition at scale,'' \emph{ICLR}, 2020.

\bibitem{liu2023mmbench}
Y.~Liu, H.~Duan, Y.~Zhang, B.~Li, S.~Zhang, W.~Zhao, Y.~Yuan, J.~Wang, C.~He, Z.~Liu, K.~Chen, and D.~Lin, ``{MMBench}: Is your multi-modal model an all-around player?'' \emph{arXiv preprint arXiv:2307.06281}, 2023.

\bibitem{ge2024openagi}
Y.~Ge, W.~Hua, K.~Mei, J.~Tan, S.~Xu, Z.~Li, Y.~Zhang \emph{et~al.}, ``{OpenAGI}: When {LLM} meets domain experts,'' in \emph{NeurIPS}, vol.~36, 2024.

\bibitem{zhu20242afc}
H.~Zhu, X.~Sui, B.~Chen, X.~Liu, P.~Chen, Y.~Fang, and S.~Wang, ``2{AFC} prompting of large multimodal models for image quality assessment,'' \emph{IEEE Trans. Circuits Syst. Video Technol.}, pp. 1--1, 2024.

\bibitem{zhao2024explainability}
H.~Zhao, H.~Chen, F.~Yang, N.~Liu, H.~Deng, H.~Cai, S.~Wang, D.~Yin, and M.~Du, ``Explainability for large language models: A survey,'' \emph{ACM Trans. Intell. Syst. Technol.}, vol.~15, no.~2, pp. 1--38, 2024.

\bibitem{lin2024detecting}
L.~Lin, N.~Gupta, Y.~Zhang, H.~Ren, C.-H. Liu, F.~Ding, X.~Wang, X.~Li, L.~Verdoliva, and S.~Hu, ``Detecting multimedia generated by large ai models: A survey,'' \emph{arXiv preprint arXiv:2402.00045}, 2024.

\bibitem{sha2023fake}
Z.~Sha, Z.~Li, N.~Yu, and Y.~Zhang, ``De-fake: Detection and attribution of fake images generated by text-to-image generation models,'' in \emph{CCS}.\hskip 1em plus 0.5em minus 0.4em\relax ACM, 2023, pp. 3418--3432.

\bibitem{wu2023cheap}
G.~Wu, W.~Wu, X.~Liu, K.~Xu, T.~Wan, and W.~Wang, ``Cheap-fake detection with llm using prompt engineering,'' in \emph{ICMEW}.\hskip 1em plus 0.5em minus 0.4em\relax IEEE, 2023, pp. 105--109.

\bibitem{chang2023antifakeprompt}
Y.-M. Chang, C.~Yeh, W.-C. Chiu, and N.~Yu, ``Antifakeprompt: Prompt-tuned vision-language models are fake image detectors,'' \emph{arXiv preprint arXiv:2310.17419}, 2023.

\bibitem{zhu2023gendet}
M.~Zhu, H.~Chen, M.~Huang, W.~Li, H.~Hu, J.~Hu, and Y.~Wang, ``Gendet: Towards good generalizations for ai-generated image detection,'' \emph{arXiv preprint arXiv:2312.08880}, 2023.

\bibitem{BorjiIVC2023}
A.~Borji, ``Qualitative failures of image generation models and their application in detecting deepfakes,'' \emph{Image Vision Comput.}, vol. 137, no.~C, 2023.

\bibitem{WolterML2023}
M.~Wolter, F.~Blanke, R.~Heese, and J.~Garcke, ``Wavelet-packets for deepfake image analysis and detection,'' \emph{Mach. Learn.}, vol. 111, pp. 4295--4327, 2022.

\bibitem{EpsteinICCVW2023}
D.~C. Epstein, I.~Jain, O.~Wang, and R.~Zhang, ``Online detection of {AI}-generated images,'' in \emph{ICCVW}.\hskip 1em plus 0.5em minus 0.4em\relax IEEE, 2023, pp. 382--392.

\bibitem{JUTMM2024}
Y.~Ju, S.~Jia, J.~Cai, H.~Guan, and S.~Lyu, ``{GLFF}: Global and local feature fusion for ai-synthesized image detection,'' \emph{IEEE Trans. Multimedia}, vol.~26, pp. 4073--4085, 2024.

\bibitem{sarkar2024shadows}
A.~Sarkar, H.~Mai, A.~Mahapatra, S.~Lazebnik, D.~A. Forsyth, and A.~Bhattad, ``Shadows don't lie and lines can't bend! generative models don't know projective geometry... for now,'' in \emph{CVPR}.\hskip 1em plus 0.5em minus 0.4em\relax IEEE, 2024, pp. 28\,140--28\,149.

\bibitem{wang2019fakespotter}
R.~Wang, F.~Juefei-Xu, L.~Ma, X.~Xie, Y.~Huang, J.~Wang, and Y.~Liu, ``Fakespotter: A simple yet robust baseline for spotting ai-synthesized fake faces,'' \emph{IJCAI}, pp. 3444--3451, 2020.

\bibitem{verdoliva2022}
L.~Verdoliva, D.~Cozzolino, and K.~Nagano, ``2022 ieee image and video processing cup synthetic image detection,'' \emph{IEEE}, 2022.

\bibitem{he2021forgerynet}
Y.~He, B.~Gan, S.~Chen, Y.~Zhou, G.~Yin, L.~Song, L.~Sheng, J.~Shao, and Z.~Liu, ``Forgerynet: A versatile benchmark for comprehensive forgery analysis,'' in \emph{CVPR}.\hskip 1em plus 0.5em minus 0.4em\relax IEEE, 2021, pp. 4360--4369.

\bibitem{lu2024seeing}
Z.~Lu, D.~Huang, L.~Bai, J.~Qu, C.~Wu, X.~Liu, and W.~Ouyang, ``Seeing is not always believing: benchmarking human and model perception of ai-generated images,'' \emph{NeurIPS}, vol.~36, 2024.

\bibitem{zhu2024genimage}
M.~Zhu, H.~Chen, Q.~Yan, X.~Huang, G.~Lin, W.~Li, Z.~Tu, H.~Hu, J.~Hu, and Y.~Wang, ``Genimage: A million-scale benchmark for detecting ai-generated image,'' in \emph{NeurIPS}, 2024.

\bibitem{wang2023benchmarking}
Y.~Wang, Z.~Huang, and X.~Hong, ``Benchmarking deepart detection,'' \emph{arXiv preprint arXiv:2302.14475}, 2023.

\bibitem{hong2024wildfake}
Y.~Hong and J.~Zhang, ``Wildfake: A large-scale challenging dataset for {AI}-generated images detection,'' \emph{arXiv preprint arXiv:2402.11843}, 2024.

\bibitem{karras2017progressive}
T.~Karras, T.~Aila, S.~Laine, and J.~Lehtinen, ``Progressive growing of {GAN}s for improved quality, stability, and variation,'' in \emph{ICLR}, 2018.

\bibitem{karras2019style}
T.~Karras, S.~Laine, and T.~Aila, ``A style-based generator architecture for generative adversarial networks,'' in \emph{CVRP}.\hskip 1em plus 0.5em minus 0.4em\relax IEEE, 2019, pp. 4401--4410.

\bibitem{ding2022cogview2}
M.~Ding, W.~Zheng, W.~Hong, and J.~Tang, ``Cogview2: Faster and better text-to-image generation via hierarchical transformers,'' in \emph{NeurIPS}, 2022, pp. 16\,890--16\,902.

\bibitem{wu2023human}
X.~Wu, Y.~Hao, K.~Sun, Y.~Chen, F.~Zhu, R.~Zhao, and H.~Li, ``Human preference score v2: A solid benchmark for evaluating human preferences of text-to-image synthesis,'' \emph{arXiv preprint arXiv:2306.09341}, 2023.

\bibitem{liu2021fusedream}
X.~Liu, C.~Gong, L.~Wu, S.~Zhang, H.~Su, and Q.~Liu, ``{FuseDream}: Training-free text-to-image generation with improved {CLIP}+{GAN} space optimization,'' \emph{arXiv preprint arXiv:2112.01573}, 2021.

\bibitem{gu2022vector}
S.~Gu, D.~Chen, J.~Bao, F.~Wen, B.~Zhang, D.~Chen, L.~Yuan, and B.~Guo, ``Vector quantized diffusion model for text-to-image synthesis,'' in \emph{CVPR}.\hskip 1em plus 0.5em minus 0.4em\relax IEEE, 2022, pp. 10\,696--10\,706.

\bibitem{nichol2021glide}
A.~Nichol, P.~Dhariwal, A.~Ramesh, P.~Shyam, P.~Mishkin, B.~McGrew, I.~Sutskever, and M.~Chen, ``Glide: Towards photorealistic image generation and editing with text-guided diffusion models,'' in \emph{ICML}.\hskip 1em plus 0.5em minus 0.4em\relax PMLR, 2022, pp. 16\,784--16\,804.

\bibitem{rombach2022high}
R.~Rombach, A.~Blattmann, D.~Lorenz, P.~Esser, and B.~Ommer, ``High-resolution image synthesis with latent diffusion models,'' in \emph{CVPR}.\hskip 1em plus 0.5em minus 0.4em\relax IEEE, 2022, pp. 10\,684--10\,695.

\bibitem{li2023agiqa}
C.~Li, Z.~Zhang, H.~Wu, W.~Sun, X.~Min, X.~Liu, G.~Zhai, and W.~Lin, ``{AGIQA-3K}: An open database for ai-generated image quality assessment,'' \emph{IEEE Trans. Circuits Syst. Video Technol.}, 2023.

\bibitem{yuan2023pku}
J.~Yuan, X.~Cao, C.~Li, F.~Yang, J.~Lin, and X.~Cao, ``{PKU-I2IQA}: An image-to-image quality assessment database for ai generated images,'' \emph{arXiv preprint arXiv:2311.15556}, 2023.

\bibitem{fu2023dreamsim}
S.~Fu, N.~Tamir, S.~Sundaram, L.~Chai, R.~Zhang, T.~Dekel, and P.~Isola, ``Dreamsim: Learning new dimensions of human visual similarity using synthetic data,'' in \emph{NeurIPS}, 2023.

\bibitem{wang2022diffusiondb}
Z.~J. Wang, E.~Montoya, D.~Munechika, H.~Yang, B.~Hoover, and D.~H. Chau, ``{DiffusionDB}: A large-scale prompt gallery dataset for text-to-image generative models,'' in \emph{ACL}, 2023, pp. 893--911.

\bibitem{ramesh2022hierarchical}
A.~Ramesh, P.~Dhariwal, A.~Nichol, C.~Chu, and M.~Chen, ``Hierarchical text-conditional image generation with {CLIP} latents,'' \emph{arXiv preprint arXiv:2204.06125}, 2022.

\bibitem{dalle3reddit}
\BIBentryALTinterwordspacing
``Dalle3 reddit dataset,'' 2024. [Online]. Available: \url{https://huggingface.co/datasets/ProGamerGov/dalle-3-reddit-dataset}
\BIBentrySTDinterwordspacing

\bibitem{midjourney-v5-dataset}
\BIBentryALTinterwordspacing
``midjourney-v5 prompt dataset,'' 2024. [Online]. Available: \url{https://huggingface.co/datasets/tarungupta83/MidJourney_v5_Prompt_dataset}
\BIBentrySTDinterwordspacing

\bibitem{zhu2023minigpt}
D.~Zhu, J.~Chen, X.~Shen, X.~Li, and M.~Elhoseiny, ``{Minigpt-4}: Enhancing vision-language understanding with advanced large language models,'' in \emph{ICLR}, 2024.

\bibitem{li2023otter}
B.~Li, Y.~Zhang, L.~Chen, J.~Wang, J.~Yang, and Z.~Liu, ``Otter: A multi-modal model with in-context instruction tuning,'' \emph{arXiv preprint arXiv:2305.03726}, 2023.

\bibitem{achiam2023gpt}
J.~Achiam, S.~Adler, S.~Agarwal, L.~Ahmad, I.~Akkaya, F.~L. Aleman, D.~Almeida, J.~Altenschmidt, S.~Altman, S.~Anadkat \emph{et~al.}, ``{GPT-4} technical report,'' \emph{arXiv preprint arXiv:2303.08774}, 2023.

\bibitem{geminiteam2024gemini}
G.~Team, ``Gemini: A family of highly capable multimodal models,'' \emph{arXiv preprint arXiv:2312.11805}, 2024.

\bibitem{claude3}
Anthrop, ``Introducing the next generation of claude,'' \url{https://www.anthropic.com/news/claude-3-family}, 2024.03.

\bibitem{li2024seed}
B.~Li, Y.~Ge, Y.~Ge, G.~Wang, R.~Wang, R.~Zhang, and Y.~Shan, ``Seed-bench: Benchmarking multimodal large language models,'' in \emph{CVPR}.\hskip 1em plus 0.5em minus 0.4em\relax IEEE, 2024, pp. 13\,299--13\,308.

\bibitem{cai2023benchlmm}
R.~Cai, Z.~Song, D.~Guan, Z.~Chen, X.~Luo, C.~Yi, and A.~Kot, ``{BenchLMM}: Benchmarking cross-style visual capability of large multimodal models,'' \emph{arXiv preprint arXiv:2312.02896}, 2023.

\bibitem{shi2023chef}
Z.~Shi, Z.~Wang, H.~Fan, Z.~Yin, L.~Sheng, Y.~Qiao, and J.~Shao, ``{ChEF}: A comprehensive evaluation framework for standardized assessment of multimodal large language models,'' \emph{arXiv preprint arXiv:2311.02692}, 2023.

\bibitem{wu2024comprehensive}
T.~Wu, K.~Ma, J.~Liang, Y.~Yang, and L.~Zhang, ``A comprehensive study of multimodal large language models for image quality assessment,'' \emph{arXiv preprint arXiv:2403.10854}, 2024.

\bibitem{huang2024aesbench}
Y.~Huang, Q.~Yuan, X.~Sheng, Z.~Yang, H.~Wu, P.~Chen, Y.~Yang, L.~Li, and W.~Lin, ``{AesBench}: An expert benchmark for multimodal large language models on image aesthetics perception,'' in \emph{ACM MM}.\hskip 1em plus 0.5em minus 0.4em\relax ACM, 2024.

\bibitem{lu2022learn}
P.~Lu, S.~Mishra, T.~Xia, L.~Qiu, K.-W. Chang, S.-C. Zhu, O.~Tafjord, P.~Clark, and A.~Kalyan, ``Learn to explain: Multimodal reasoning via thought chains for science question answering,'' in \emph{NeurIPS}, 2022, pp. 2507--2521.

\bibitem{talmor2018commonsenseqa}
A.~Talmor, J.~Herzig, N.~Lourie, and J.~Berant, ``{CommonsenseQA}: A question answering challenge targeting commonsense knowledge,'' in \emph{ACL}, 2018, pp. 4149--4158.

\bibitem{deng2009imagenet}
J.~Deng, W.~Dong, R.~Socher, L.-J. Li, K.~Li, and L.~Fei-Fei, ``{ImageNet}: A large-scale hierarchical image database,'' in \emph{CVPR}.\hskip 1em plus 0.5em minus 0.4em\relax IEEE, 2009, pp. 248--255.

\bibitem{agustsson2017ntire}
E.~Agustsson and R.~Timofte, ``Ntire 2017 challenge on single image super-resolution: Dataset and study,'' in \emph{CVPRW}.\hskip 1em plus 0.5em minus 0.4em\relax IEEE, 2017, pp. 126--135.

\bibitem{park2024can}
H.~Park, G.~Kim, D.~Lee, and H.~K. Kim, ``Can you spot the ai-generated images? distinguishing fake images using signal detection theory,'' in \emph{HCI International}.\hskip 1em plus 0.5em minus 0.4em\relax Springer, 2024, pp. 299--313.

\bibitem{chen2023exploring}
Z.~Chen, W.~Sun, H.~Wu, Z.~Zhang, J.~Jia, X.~Min, G.~Zhai, and W.~Zhang, ``Exploring the naturalness of ai-generated images,'' \emph{arXiv preprint arXiv:2312.05476}, 2023.

\bibitem{farid2022lighting}
H.~Farid, ``Lighting (in) consistency of paint by text,'' \emph{arXiv preprint arXiv:2207.13744}, 2022.

\bibitem{farid2022perspective}
------, ``Perspective (in) consistency of paint by text,'' \emph{arXiv preprint arXiv:2206.14617}, 2022.

\bibitem{rubin2021learning}
O.~Rubin, J.~Herzig, and J.~Berant, ``Learning to retrieve prompts for in-context learning,'' in \emph{ACL}, 2022, pp. 2655--2671.

\bibitem{WeiNIPS2022}
J.~Wei, X.~Wang, D.~Schuurmans, M.~Bosma, b.~ichter, F.~Xia, E.~Chi, Q.~V. Le, and D.~Zhou, ``Chain-of-thought prompting elicits reasoning in large language models,'' in \emph{NeurIPS}, 2022, pp. 24\,824--24\,837.

\bibitem{wu2022survey}
X.~Wu, L.~Xiao, Y.~Sun, J.~Zhang, T.~Ma, and L.~He, ``A survey of human-in-the-loop for machine learning,'' \emph{Future Gener. Comp. Sy.}, vol. 135, pp. 364--381, 2022.

\bibitem{ye2023mplugowl2}
Q.~Ye, H.~Xu, J.~Ye, M.~Yan, A.~Hu, H.~Liu, Q.~Qian, J.~Zhang, F.~Huang, and J.~Zhou, ``mplug-owl2: Revolutionizing multi-modal large language model with modality collaboration,'' \emph{arXiv preprint arXiv:2311.04257}, 2023.

\bibitem{wu2024q}
H.~Wu, Z.~Zhang, E.~Zhang, C.~Chen, L.~Liao, A.~Wang, K.~Xu, C.~Li, J.~Hou, G.~Zhai \emph{et~al.}, ``Q-{I}nstruct: {I}mproving low-level visual abilities for multi-modality foundation models,'' in \emph{Proceedings of the IEEE/CVF Conference on Computer Vision and Pattern Recognition}, 2024, pp. 25\,490--25\,500.

\bibitem{du2022glm}
Z.~Du, Y.~Qian, X.~Liu, M.~Ding, J.~Qiu, Z.~Yang, and J.~Tang, ``{GLM}: General language model pretraining with autoregressive blank infilling,'' in \emph{ACL}, 2022, pp. 320--335.

\bibitem{laurençon2023idefics}
H.~Laurençon, L.~Saulnier, L.~Tronchon, S.~Bekman, A.~Singh, A.~Lozhkov, T.~Wang, S.~Karamcheti, A.~M. Rush, D.~Kiela, M.~Cord, and V.~Sanh, ``{OBELICS}: An open web-scale filtered dataset of interleaved image-text documents,'' in \emph{NeurIPS}, 2023.

\bibitem{peng2023kosmos2}
Z.~Peng, W.~Wang, L.~Dong, Y.~Hao, S.~Huang, S.~Ma, and F.~Wei, ``Kosmos-2: Grounding multimodal large language models to the world,'' \emph{arXiv preprint arXiv: 2306.14824}, 2023.

\bibitem{dong2024internlmxcomposer2}
X.~Dong, P.~Zhang, Y.~Zang, Y.~Cao, B.~Wang, L.~Ouyang, X.~Wei, S.~Zhang, H.~Duan, M.~Cao, W.~Zhang, Y.~Li, H.~Yan, Y.~Gao, X.~Zhang, W.~Li, J.~Li, K.~Chen, C.~He, X.~Zhang, Y.~Qiao, D.~Lin, and J.~Wang, ``Internlm-xcomposer2: Mastering free-form text-image composition and comprehension in vision-language large model,'' \emph{arXiv preprint arXiv:2401.16420}, 2024.

\bibitem{yu2015lsun}
F.~Yu, A.~Seff, Y.~Zhang, S.~Song, T.~Funkhouser, and J.~Xiao, ``{LSUN}: Construction of a large-scale image dataset using deep learning with humans in the loop,'' \emph{arXiv preprint arXiv:1506.03365}, 2015.

\bibitem{chiang2023can}
C.-H. Chiang and H.-y. Lee, ``Can large language models be an alternative to human evaluations?'' \emph{arXiv preprint arXiv:2305.01937}, 2023.

\bibitem{zheng2024judging}
L.~Zheng, W.-L. Chiang, Y.~Sheng, S.~Zhuang, Z.~Wu, Y.~Zhuang, Z.~Lin, Z.~Li, D.~Li, E.~Xing \emph{et~al.}, ``Judging {LLM}-as-a-judge with {MT}-bench and chatbot arena,'' in \emph{NeurIPS}, vol.~36, 2024.

\bibitem{zhang2023multimodal}
Z.~Zhang, A.~Zhang, M.~Li, H.~Zhao, G.~Karypis, and A.~Smola, ``Multimodal chain-of-thought reasoning in language models,'' \emph{Trans. Mach. Learn. Res.}, vol. 2024, 2024.

\bibitem{fu2023chain}
Y.~Fu, L.~Ou, M.~Chen, Y.~Wan, H.~Peng, and T.~Khot, ``Chain-of-thought hub: A continuous effort to measure large language models' reasoning performance,'' \emph{arXiv preprint arXiv:2305.17306}, 2023.

\bibitem{wu2023q}
H.~Wu, Z.~Zhang, E.~Zhang, C.~Chen, L.~Liao, A.~Wang, C.~Li, W.~Sun, Q.~Yan, G.~Zhai \emph{et~al.}, ``Q-bench: A benchmark for general-purpose foundation models on low-level vision,'' in \emph{ICLR}, 2024.

\bibitem{yu2023mm}
W.~Yu, Z.~Yang, L.~Li, J.~Wang, K.~Lin, Z.~Liu, X.~Wang, and L.~Wang, ``Mm-vet: Evaluating large multimodal models for integrated capabilities,'' \emph{arXiv preprint arXiv:2308.02490}, 2023.

\bibitem{papineni2002bleu}
K.~Papineni, S.~Roukos, T.~Ward, and W.-J. Zhu, ``Bleu: A method for automatic evaluation of machine translation,'' in \emph{ACL}, 2002, pp. 311--318.

\bibitem{lin2004rouge}
C.-Y. Lin, ``Rouge: A package for automatic evaluation of summaries,'' in \emph{ACL}, 2004, pp. 74--81.

\bibitem{reimers2019sentence}
N.~Reimers and I.~Gurevych, ``{Sentence-BERT}: Sentence embeddings using siamese bert-networks,'' in \emph{EMNLP}, 2019, pp. 3982--3992.

\bibitem{zhang2022automatic}
Z.~Zhang, A.~Zhang, M.~Li, and A.~Smola, ``Automatic chain of thought prompting in large language models,'' in \emph{EMNLP}, 2022, pp. 12\,113--12\,139.

\end{thebibliography}
\end{document}